\documentclass{article}

\usepackage{microtype}
\usepackage{graphicx}
\usepackage{subfigure}
\usepackage{booktabs} 

\usepackage{hyperref}

\usepackage[accepted]{icml2024}

\usepackage{amsmath}
\usepackage{amssymb}
\usepackage{mathtools}
\usepackage{amsthm}

\usepackage[capitalize,noabbrev]{cleveref}

\theoremstyle{plain}
\newtheorem{theorem}{Theorem}[section]

\theoremstyle{definition}
\newtheorem{definition}[theorem]{Definition}

\theoremstyle{remark}
\newtheorem{remark}[theorem]{Remark}

\usepackage[textsize=tiny]{todonotes}

\usepackage{amsmath,amsfonts,bm}

\def\eqref#1{equation~\ref{#1}}

\def\1{\bm{1}}

\DeclareMathAlphabet{\mathsfit}{\encodingdefault}{\sfdefault}{m}{sl}
\SetMathAlphabet{\mathsfit}{bold}{\encodingdefault}{\sfdefault}{bx}{n}

\newcommand{\R}{\mathbb{R}}

\usepackage{caption}
\usepackage{algorithm}
\usepackage{algorithmicx}
\usepackage[noend]{algpseudocode}
\usepackage{amsmath, amsfonts, amssymb, amsthm, amsbsy, amscd, bm, bbm,mathrsfs}    \usepackage{graphicx}
\graphicspath{{figs/}}
\usepackage{cleveref}
\usepackage{enumitem}
\usepackage{tikz}

\theoremstyle{definition}

\setlist{leftmargin=5mm}
\usepackage{titlesec}
\usepackage{wrapfig}
\usepackage{multirow}

\usepackage{stfloats}
\newcommand{\x}{\bm{x}}
\renewcommand{\a}{\bm{a}}
\newcommand{\s}{\bm{s}}

\newcommand{\g}{\bm{g}}

\renewcommand{\b}{\mathbf{b}}
\newcommand{\I}{\mathbf{I}}

\newcommand{\W}{\mathbf{W}}
\renewcommand{\s}{\bm{s}}

\newcommand{\G}{\mathbf{G}}

\usepackage{pifont}
\newcommand{\cmark}{\ding{51}}%
\newcommand{\xmark}{\ding{55}}%

\colorlet{pink}{red!40}
\colorlet{betterblue}{cyan!60}

\usepackage{adjustbox}

\usepackage[most]{tcolorbox}
\newtcolorbox[auto counter]{summary}[1][]{title={\bfseries Challenge~\thetcbcounter},enhanced,drop shadow={black!50!white},
  coltitle=black,
  top=0.3in,
  attach boxed title to top left=
  {xshift=1.5em,yshift=-\tcboxedtitleheight/2},
  boxed title style={size=small,colback=pink},#1}

\icmltitlerunning{Differentially Private Bias-Term Fine-tuning of Foundation Models}

\begin{document}

\twocolumn[
\icmltitle{Differentially Private Bias-Term Fine-tuning of Foundation Models}



\icmlsetsymbol{equal}{*}

\begin{icmlauthorlist}
\icmlauthor{Zhiqi Bu}{yyy}
\icmlauthor{Yu-xiang Wang}{yyy,comp}
\icmlauthor{Sheng Zha}{yyy}
\icmlauthor{George Karypis}{yyy}
\end{icmlauthorlist}

\icmlaffiliation{yyy}
{Amazon AI}
\icmlaffiliation{comp}{University of California, San Diego}

\icmlcorrespondingauthor{Zhiqi Bu}{zhiqibu@amazon.com}

\icmlkeywords{Machine Learning, ICML}

\vskip 0.3in
]



\printAffiliationsAndNotice{}  

\begin{abstract}
We study the problem of differentially private (DP) fine-tuning of large pre-trained models -- a recent privacy-preserving approach suitable for solving downstream tasks with sensitive data. Existing work has demonstrated that high accuracy is possible under strong privacy constraint, yet requires significant computational overhead or modifications to the network architecture.
We propose differentially private bias-term fine-tuning (DP-BiTFiT), which matches the state-of-the-art accuracy for DP algorithms and the efficiency of the standard BiTFiT. DP-BiTFiT is model agnostic (not modifying the network architecture), parameter efficient (only training about $0.1\%$ of the parameters), and computation efficient (almost removing the overhead caused by DP, in both the time and space complexity). On a wide range of tasks, DP-BiTFiT is $2\sim 30\times$ faster and uses $2\sim 8\times$ less memory than DP full fine-tuning, even faster than the standard full fine-tuning. This amazing efficiency enables us to conduct DP fine-tuning on language and vision tasks with long-sequence texts and high-resolution images, which were computationally difficult using existing methods. We open-source our code at FastDP (\url{https://github.com/awslabs/fast-differential-privacy}).
\end{abstract}

\begin{figure*}[!htb]
\vspace{-0.1cm}
    \centering
    \includegraphics[width=0.33\linewidth]{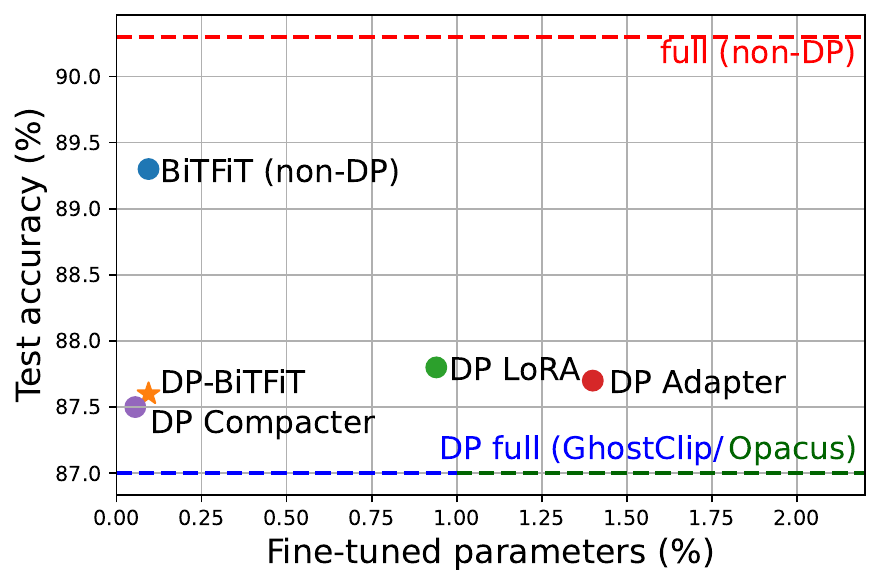}
    \hspace{-0.25cm}
    \includegraphics[width=0.33\linewidth]{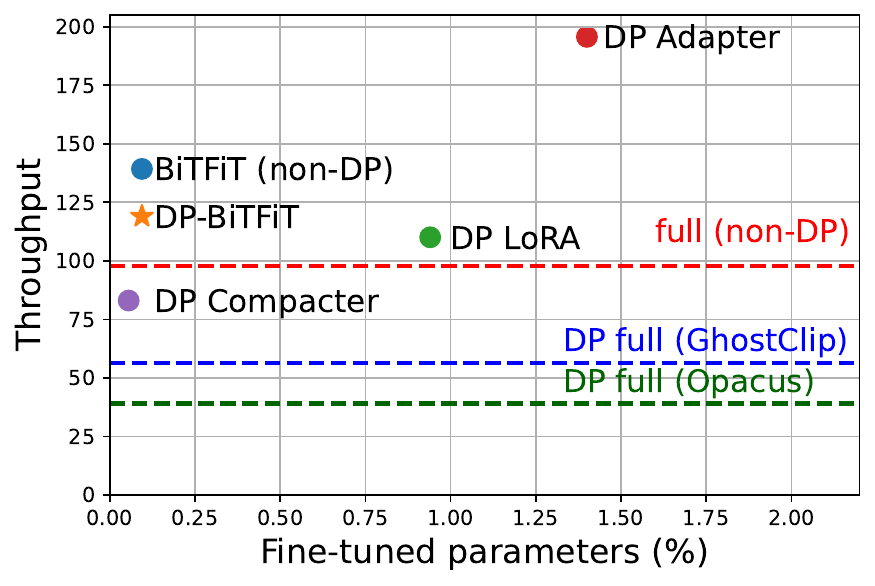}
    \hspace{-0.25cm}
    \includegraphics[width=0.33\linewidth]{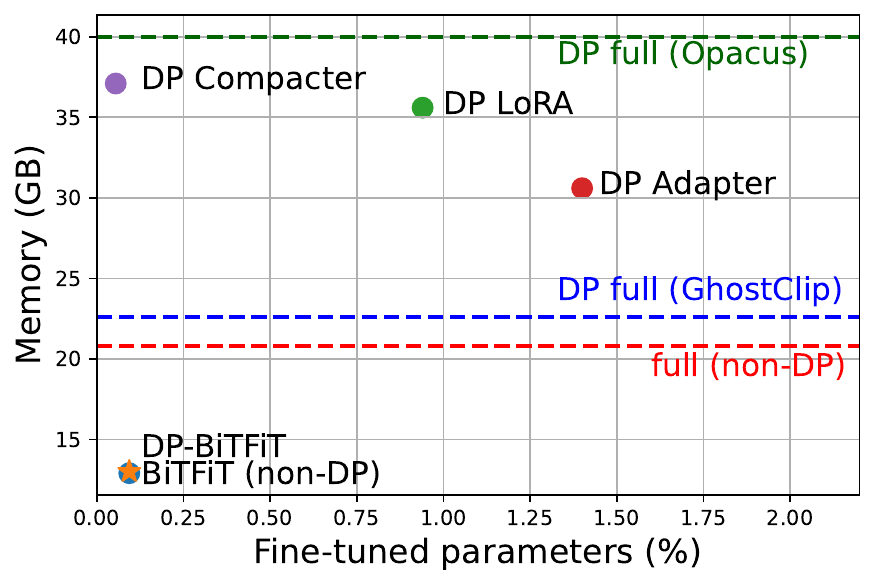}
    \vspace{-0.3cm}
\caption{Performance of different fine-tuning methods on MNLI dataset with RoBERTa-large. DP-BiTFiT is one of the most accurate (below DP LoRA marginally), fastest (only slower than DP Adapter), and memory efficient (outperforming others substantially by $3\times$) DP methods.}
\label{fig:all param efficient}
\vspace{-0.1cm}
\end{figure*}

\section{Introduction}
Fine-tuning large pre-trained neural networks is one of the most critical technique in deep learning, yielding strong performance in a variety of domains \citep{pan2009survey,kenton2019bert,goyal2017accurate}. Among different methods, full fine-tuning is the most prevalent one, which trains all the model parameters on the downstream tasks and achieves high accuracy within a small number of training epochs. 
However, full fine-tuning on large models, from hundreds of millions \citep{he2016deep,chen2016training} to billions of parameters \citep{brown2020language}, can be burdensome in terms of the computation and the deployment, since a full copy of fine-tuned model parameters is needed for each task.

To alleviate this issue, the parameter efficient fine-tuning only trains a substantially small portion of the model parameters, in contrast to the full fine-tuning. At a high level, the parameter efficient fine-tuning methods can be divided into two categories. $\langle 1\rangle $ Model-aware methods, meaning a relatively small number of parameters are introduced into the neural network architecture and only the new parameters are optimized. Examples include LoRA \citep{hu2021lora}, Adapter \citep{houlsby2019parameter}, and Compacter \citep{mahabadi2021compacter}. $\langle 2\rangle $ Model-agnostic methods, meaning that only a subset of existing parameters are trainable. Examples include training only the output linear layer (linear probing, \citep{kornblith2019better}), only the layer normalization layer \citep{houlsby2019parameter} and bias-term fine-tuning (BiTFiT) \citep{zaken2022bitfit}. We illustrate the differences as follows: $\W_0,\b_0$ are the pre-trained weights and biases, $\hat\quad$ indicates trainable parameters, and $\bm\theta$ is the additional parameters.

\vspace{-0.5cm}
\begin{align*}
\underbrace{f(\x;\W_0,\b_0)}_\text{pre-trained model}\to
\begin{cases}
{f(\x;\hat\W,\hat\b)}&\text{full fine-tuning} \\ {f(\x;\W_0,\b_0,\hat{\bm\theta})} &\text{model-aware} \\ {f(\x;\W_0,\hat\b)} &\text{bias-term only}
\end{cases}
\end{align*}

Empirically, these parameter efficient fine-tuning methods have achieved high accuracy that is comparable to full fine-tuning in the standard non-private setting. For instance, linear probing of ResNet \citep{he2016deep} and Vision Transformer (ViT, \citep{dosovitskiy2020image}) achieves 80\% accuracy on the ImageNet dataset \citep{sun2017revisiting,kornblith2019better}; LoRA and BiTFiT of RoBERTa \citep{liu2019roberta} and BERT \citep{kenton2019bert} achieve about 94\% on SST2 and on average 85\% across the General Language Understanding Evaluation (GLUE) datasets \citep{he2021towards,hu2021lora}. In addition, parameter efficient methods are faster than full fine-tuning and save the communication cost significantly in distributed learning.

Parallel to these developments, the success of deep learning models relies on the availability of large datasets, which may contain sensitive information to be protected rigorously. This privacy issue is well-known for neural networks can be vulnerable to privacy attacks: membership information can be leaked from the purchase records via Google and Amazon online services \citep{shokri2017membership}; sensitive texts can be reconstructed by specifically designed prefix on GPT2 \citep{carlini2021extracting} and so can images in CIFAR10 and MNIST \citep{haim2022reconstructing}. To protect against such privacy risks, the standard technique is differential privacy (DP, formally stated in \Cref{def:DP}) which randomizes the standard optimizers via the private gradient in \Cref{eq: private gradient}. 

A recent line of work has extensively studied the DP fine-tuning in both computer vision and language tasks, often achieving less than 3\% accuracy drop across different settings via full fine-tuning \citep{de2022unlocking,li2021large,bu2022automatic,bu2022scalable}, linear probing \citep{mehta2022large}, LoRA, Adapter, or Compacter \citep{yu2021differentially}. In fact, fine-tuning or pre-training from large dataset is considered necessary in the DP deep learning literature. As a matter of fact, full fine-tuning DP-GPT2 only achieves 24.2 BLEU score ($\epsilon=8$) on E2E dataset if randomly initialized \citep{li2021large}, in starking contrast to 63.2 BLEU if pre-trained; similarly, state-of-the-art (SOTA) DP accuracy on ImageNet is 48\% ($\epsilon=10$) without pre-training \citep{kurakin2022toward} but 86.7\% accuracy if pre-trained \citep{de2022unlocking}. Specifically, parameter efficient DP fine-tuning has empirically demonstrated strong accuracy (see our \Cref{tab:finetuning comparison}) with $3\sim 4\times$ memory saving and $2\sim 3\times$ speedup compared to DP full fine-tuning by Opacus \citep[c.f. \Cref{fig:memory_time_vs_sequence_roberta} and ][Table 3]{yu2021differentially}. Although previous works have shed light on various DP fine-tuning methods, we are the first to study DP-BiTFiT specifically and to show two distinctive advantages of it.

Firstly, DP-BiTFiT is model-agnostic and remains its parameter efficiency around 0.1\% across models by \Cref{tab:parameter efficiency}. While linear probing is also model-agnostic, the parameter efficiency can be as high as 8\% in ResNet50. Other methods like LoRA, Adapter and Compacter are architecture-dependent and possibly parameter inefficient, making them difficult to directly apply on arbitrary neural networks: LoRA and Adapter may need to train more than 12\% on BART-large \citep{lewis2020bart} to achieve high accuracy by \citet[Figure 1\& 4]{he2021towards}.

Secondly, DP-BiTFiT is computationally efficient, almost as much as the standard BiTFiT and significantly more efficient than DP full fine-tuning, particularly with large models and high-dimensional input data. For examples of DP full fine-tuning, \citet{li2021large} have reported $2\sim 4\times$ slowdown on large language models for four advanced private codebases and up to $5\times$ memory overhead, compared to the standard fine-tuning; even on small networks,  11 codebases  across Tensorflow, JAX, and Pytorch have demonstrated $0.2\sim 5\times$ slowdown and $3\sim 100\times$ reduction in maximum batch size in \citet{subramani2021enabling}. See more discussion in \Cref{sec:scalability of DP}.

\vspace{-0.2cm}
\begin{algorithm}[H]
\caption{\colorbox{pink}{DP} Bias-Term Fine-Tuning (BiTFiT)}
\label{alg:dp bitfit}
\textbf{Parameters:} $l$-th layer's bias $\b_l$, subsampling probability $p$, number of iterations $T$, number of layers $L$, noise scale $\sigma$, clipping threshold $R$, clipping factor $C_i$ (if no clipping then $C_i=1$).
\begin{algorithmic}[1]
    \For{iteration $t = 1, \cdots, T$}
			\State{Subsample a batch $B_t\subseteq \{1, \ldots, n\}$ from training set with probability $p$}
			
			\For{layer $l\in L,L-1,\cdots,1$}
			\State Get output gradient $\frac{\partial \mathcal{L}}{\partial \s_l}$
			\State {Compute per-example gradient and its norm:}
        \hspace{1cm}\State{$\frac{\partial \mathcal{L}_i}{\partial\b_{l}}=\frac{\partial \mathcal{L}}{\partial \s_{l,i}}^\top \mathbf{1}$
			\colorbox{pink}{$\Longrightarrow\|\frac{\partial \mathcal{L}_i}{\partial\b_{l}}\|_F^2$}
			}
			\EndFor
			
			\State \colorbox{pink}{Aggregate grad norms: $\|\frac{\partial \mathcal{L}_i}{\partial\b}\|_F^2=\sum_l\|\frac{\partial \mathcal{L}_i}{\partial\b_{l}}\|_F^2$}
			\State \colorbox{pink}{Compute clipping factor: $C_{i}=C(\|\frac{\partial \mathcal{L}_i}{\partial\b}\|_F;R)$}
			\State Compute sum of clipped gradients $\G=\sum_i C_i\frac{\partial \mathcal{L}_i}{\partial \b}$
			\State \colorbox{pink}{Add Gaussian noise $\G=\G+\sigma R\cdot \mathcal{N}(0, \I)$}
			\State Descend on bias terms with $\G$ by SGD/Adam/...
			\EndFor
		\end{algorithmic}
	\end{algorithm}

\noindent\textbf{Contributions.} We develop DP-BiTFiT, a fine-tuning method that is model-agnostic, accurate, privacy-preserving, parameter efficient, and computationally efficient.

\begin{enumerate}
	\item Algorithmically, we propose the Differentially Private Bias-Term Fine-Tuning (\textbf{DP-BiTFiT}) in \Cref{alg:dp bitfit} that is highly accurate under DP constraint, on par with SOTA in \Cref{sec:experiments} and even outperforming fully fine-tuned GPT2-large. 
	\item DP-BiTFiT is \textit{\textbf{model-agnostic}}\footnote{In \Cref{sec:experiments}, DP-BiTFiT is applicable to all model architectures tested, unlike LoRA (mostly only applies to transformers) and last-layer training (mostly only works on vision models).} and only optimizes 0.1\% of the model parameters on BERT, RoBERTa, GPT2, ViT, ResNet, and so on (see \Cref{tab:parameter efficiency}). Thus DP-BiTFiT is one of the \textit{\textbf{most parameter efficient}} fine-tuning methods among DP LoRA, Adapter, last-layer, etc.
	\item We design a \textit{\textbf{computationally efficient}} implementation of DP-BiTFiT, whose time and space complexity is almost the same as the standard non-DP BiTFiT, while being faster than non-DP full fine-tuning and other DP fine-tuning (see \Cref{fig:all param efficient}). This advantage is analyzed in \Cref{tab:complexity overhead weight bias}, and demonstrated via the substantial speedup and memory-saving in \Cref{fig:memory_time_vs_sequence_roberta} and \Cref{fig:memory_time_vs_models}.
	\item DP-BiTFiT is a unique algorithm in that \textit{\textbf{the computation overhead is independent of the feature dimension}} $T$\footnote{The computation overhead to get the per-sample weight gradient norm is linear (by instantiating per-sample gradints) or quadratic in $T$ (if using the ghost norm trick \citep{goodfellow2015efficient,li2021large}), for DP full and any other PEFT.} (see \textcolor{red}{red texts} in \Cref{tab:complexity overhead weight bias}). This is due to the \textit{\textbf{activation-free forward pass}} that only happens in the no-weight training\footnote{We distinguish the weight training and bias training in \Cref{sec:backward} using the chain rules. Note that activation-free means memory-saving, which is not leveraged by DP full, LoRA, Adapter, Compacter, etc.} unlike LoRA. In \Cref{fig:all param efficient}, although DP-BiTFiT optimizes a similar number of parameters to DP LoRA or Compacter, its memory efficiency is dominating. Therefore, DP-BiTFiT enjoys a special advantage on long-sequence texts and high-resolution images (see \Cref{fig:memory_time_vs_sequence_roberta}).
\end{enumerate}

\noindent\textbf{Novelty.} At a glance, our results may appear to be incremental as we are merely adding differential privacy to an existing method (BiTFiT) through a standard mechanism (DP-SGD). This is not true! Computationally, our implementation of DP-BiTFiT is distinct and orthogonal to existing DP algorithms such as GhostClip \cite{li2021large}\footnote{Ghost clipping (GhostClip) is an algebraic technique that only works on weight gradients because it manipulates the activation tensors at $O(BT^2)$ cost. This is too expensive for high-dimension features, hence not applicable to the bias gradients.}, in that DP-BiTFiT exploits the special structures in the forward and backward passes (see the simplicity of computation graph in \Cref{fig:flowcharts}), hence removing the computational and memory overhead in DP-SGD (see the independence of $T$ in \Cref{tab:complexity overhead weight bias}), which is unavoidable in other methods. 

Our main contributions also include 
\begin{itemize}
\item The complexity analysis of DP parameter-efficient fine-tuning (PEFT) in \Cref{tab:complexity overhead weight bias} and \Cref{tab:complexity overhead weight bias param eff}. This was a missing piece in previous DP and non-DP PEFT literature (including the BiTFiT paper) and significantly helpful in determining the benefit of applying different PEFT methods. Specifically, we leverage the complexity analysis to rigorously show that the complexity saving of DP-BiTFiT is 50\% compared to the full fine-tuning, and to reveal the unique benefit of DP-BiTFiT on high-dimension data.
\item The engineering effort: at the time of writing this paper, none of existing codebases including GhostClip and Opacus remove the forward hooks, because no analysis has established that only BiTFiT can be activation-free, not LoRA/Adapter/Compactor or full fine-tuning. Our algorithm enables DP-BiTFiT by one line of code\footnote{In Pytorch, DP-BiTFiT can be enabled within our codebase by \texttt{[param.requires\_grad\_(0) for name,param in
model.named\_parameters() if 'bias' in name]}.}.
\end{itemize}

\section{Preliminaries}
\noindent\textbf{Fine-tuning methods.}
Fine-tuning, i.e. training a model on a large dataset for a sufficiently long time, and then continuing to train (or transferring) onto the downstream datasets, is the standard paradigm to achieve high accuracy in both the standard and the DP regimes. In DP deep learning, the pre-training takes place on a public dataset using regular optimizers like SGD, and the fine-tuning takes place on a private dataset which requires privacy protection, using DP optimizers like DP-SGD in \Cref{sec:dp prelim}.

In a long line of research, various fine-tuning methods have been proposed. One of the most popular method is the full fine-tuning, which simply runs gradient descents on all trainable weights and biases, thus can be inefficient when the model is large. To improve the efficiency, \citet{li2021prefix} proposes the prefix tuning that only optimizes the prompts or the input layer activation \citep{lester2021power,liu2021gpt}. However, as pointed out in \citet{hu2021lora} and \citet{li2021large}, the prefix tuning can be difficult to optimize and thus sub-optimal on large models. Another approach is to reduce the number of trainable parameters. For example, LoRA \citep{hu2021lora}, Adapter \citep{houlsby2019parameter,rebuffi2017learning,pfeiffer2021adapterfusion,ruckle2021adapterdrop,lin2020exploring} and Compacter \citep{mahabadi2021compacter} insert small `adapter' layers (usually 1-10\% of total parameters) between existing layers, and only the newly added adapters are optimized. We describe the forms and complexity of LoRA and Adapter in \Cref{app:param eff and complexity}.

In addition to the aforementioned methods, BiTFiT is a special parameter-efficient method that rivals the full fine-tuning \citep{zaken2022bitfit,cai2020tinytl,he2021towards}. Firstly, BiTFiT optimizes a subset of original parameters -- the bias terms, which usually constitute less than 1/1000 of all parameters as demonstrated in \Cref{tab:parameter efficiency}. Therefore, BiTFiT can be readily deployed to any network in a model-agnostic manner. Secondly, BiTFiT is fundamentally different to other parameter efficient methods such as LoRA, since the bias gradients are computed differently than the weight gradients on the computation graph. We will elaborate on this in \Cref{eq:outer product linear}.

\label{sec:dp prelim}
\noindent\textbf{Deep learning with differential privacy.}
We recall the classic $(\epsilon,\delta)$-DP, under which we train deep neural networks with provably privacy guarantees.
\begin{definition}[\citep{dwork2006calibrating}]\label{def:DP}
A randomized algorithm $M$ is $ (\varepsilon, \delta)$-differentially private if, for any two neighboring datasets $S,S^{\prime}$ that differ by one datapoint and for any event $E$, we have
$ \mathbb{P}[M(S) \in E] \leqslant \mathrm{e}^{\varepsilon} \mathbb{P}\left[M\left(S^{\prime}\right) \in E\right]+\delta.
$
\end{definition}
In deep learning, DP can be achieved through applying an off-the-shelf optimizer (SGD or Adam) with a privately released stochastic gradient in place of the regular $\sum_i \g_i$. The private stochastic gradient is computed by first getting a minibatch $\mathcal{I}$ via Poisson sampling, then compute
\begin{align}
	\text{Private gradient:}&\sum\limits_{i\in\mathcal{I}} \g_i\cdot C(\|\g_i\|;R)+\sigma R\cdot\mathcal{N}(0,\I)
	\label{eq: private gradient}
\end{align}
where $C$ is any function\footnote{Examples of gradient clipping include but not limited to Abadi's clipping $\min(R/\|\g_i\|,1)$ \citep{abadi2016deep} and automatic clipping (AUTO-S) $R/(\|\g_i\|+0.01)$ \citep{bu2022automatic,yang2022normalized}.} $\R^+\to\R$ subject to $C(x)\leq R/x$, $\g_i$ is the $i$-th per-sample gradient, $R$ is the clipping threshold, and $\sigma$ is the noise multiplier. The private gradient is guaranteed to be DP through the \emph{sampled-Gaussian mechanism} and the associated tight privacy accounting to compose over the iterations \citep[see, e.g.,][and the references therein.]{abadi2016deep,wang2019subsampled,mironov2019sampled,koskela2020computing,bu2020deep,gopi2021numerical}.

\label{sec:backward}
\noindent\textbf{Backward propagation.}
We briefly introduce the back-propagation, which reveals a simple yet important difference between the gradients of weights and those of biases. We consider a linear layer, indexed as  the $l$-th layer, with weight $\W_l \in \R^{d\times p}$ and bias as $\b_l \in \R^{p}$. We leave the derivation of other layers such as normalization and convolution in \Cref{app:detail back-propagation}. We denote the mini-batched input of this layer as $\a_l \in \R^{B\times T\times d}$ and the immediate output as $\s_l \in \R^{B\times T\times p}$, where $B$ is the batch size and $T$ is the feature dimension\footnote{In sequential data such as text, $T$ is the sequence length; in vision data, $T$ is the product of input dimensions (e.g. for images, $T$ is the product of height and width). We refer to a high-dimensional input when $T$ is large.\label{footnote3}}:
$\a_{l+1} = \phi (\s_l),  \s_{l}=\a_{l}\W_{l}+\b_{l}$.
Here $\phi$ is any non-parametric inter-layer operation, e.g. the non-linear activation (like ReLU), pooling, padding, and so on.

We write $\mathcal{L}=\sum_{i=1}^n \mathcal{L}_i$ as the total loss  and $\mathcal{L}_i$ as the per-sample loss of the $i$-th sample. During a standard back-propagation of $L$ layers, the chain rule keeps track of the \textit{output gradient} at each layer in a just-in-time fashion:
\begin{align}
\begin{split}
\frac{\partial \mathcal{L}}{\partial \s_l}
=&
\frac{\partial \mathcal{L}}{\partial \a_L}\circ \frac{\partial \a_{L}}{\partial \s_{L-1}}\cdot\frac{\partial \s_{L-1}}{\partial \a_{L-1}}\circ\cdots\frac{\partial \a_{l+1}}{\partial \s_{l}}
\\
=& \frac{\partial \mathcal{L}}{\partial \s_{l+1}}\W_{l+1}\circ\phi^{\prime}(\s_{l}).
\end{split}
\label{eq:back prop1}
\end{align}
Here $\circ$ is the Hadamard product and $\cdot$ is the matrix product. 
This output gradient $\frac{\partial\mathcal{L}}{\partial \s_l}$ is used to compute per-sample gradient of weights and biases,
\begin{align}
\begin{split}
\frac{\partial \mathcal{L}_i}{\partial \W_{l}}^\top=\sum_j\frac{\partial \mathcal{L}_i}{\partial \s_{l,j}}^\top\frac{\partial \s_{l,j}}{\partial \W_{l}}=\frac{\partial \mathcal{L}}{\partial \s_{l,i}}^\top\a_{l,i}, 
\\
\frac{\partial \mathcal{L}_i}{\partial \b_{l}}^\top=\sum_j\frac{\partial \mathcal{L}_i}{\partial \s_{l,j}}^\top\frac{\partial \s_{l,j}}{\partial \b_{l}}=\frac{\partial \mathcal{L}}{\partial \s_{l,i}}^\top\mathbf{1}.
\end{split}
\label{eq:outer product linear}
\end{align}
Notably, the weight gradient needs the activation tensor $\a_l$ to compute an expensive $O(BTpd)$ tensor multiplication. Memory-wise, $\{\a_l\}_l$ across all layers is very costly to store (taking more than 95\% memory across VGG, ResNet, DenseNet, RoBERTa, etc. by \citet[Figure 3]{jain2020checkmate}). In sharp contrast, the computation of bias gradient does not need $\a_l$, and the multiplication with $\mathbf{1}$ in \Cref{eq:outer product linear} is actually a cheap $O(BTp)$ summation on $\frac{\partial \mathcal{L}}{\partial \s_l}: B\times T\times p\to B\times p$.

\label{sec:forward}
\noindent\textbf{Forward propagation and the hook.}
During the forward propagation, all Pytorch-based codebases for DP algorithms such as Private Transformers, Opacus, FastGradClip, Private-Vision, and others \citep{yu2021differentially,bu2023differentially} register the forward hooks to extract the activation tensors $\{\a_l\}_l$ of all layers from the computation graph, where $\a_l$ is computed and stored. Hence, the majority of memory burden is on the activation that grows extremely large for huge models like GPT3 \citep{brown2020language} with 175B parameters: the activation tensors consume more than 3600GB of memory while the parameters and gradients only consume 300GB \citep{rajbhandari2020zero}. On one hand, this issue can be alleviated by the activation recomputation or checkpointing technique \citep{chen2016training,jain2020checkmate}, whose memory cost reduces from $O(L)$ to $O(\sqrt{L})$ with an extra 33\% slowdown. Alternatively, we note that the activation tensors are not necessary in the forward propagation, if we only optimize the bias terms.

\section{Differentially private Bias-Term Fine-Tuning}
\begin{figure*}[!htb]
	\centering
	\includegraphics[width=0.95\linewidth]{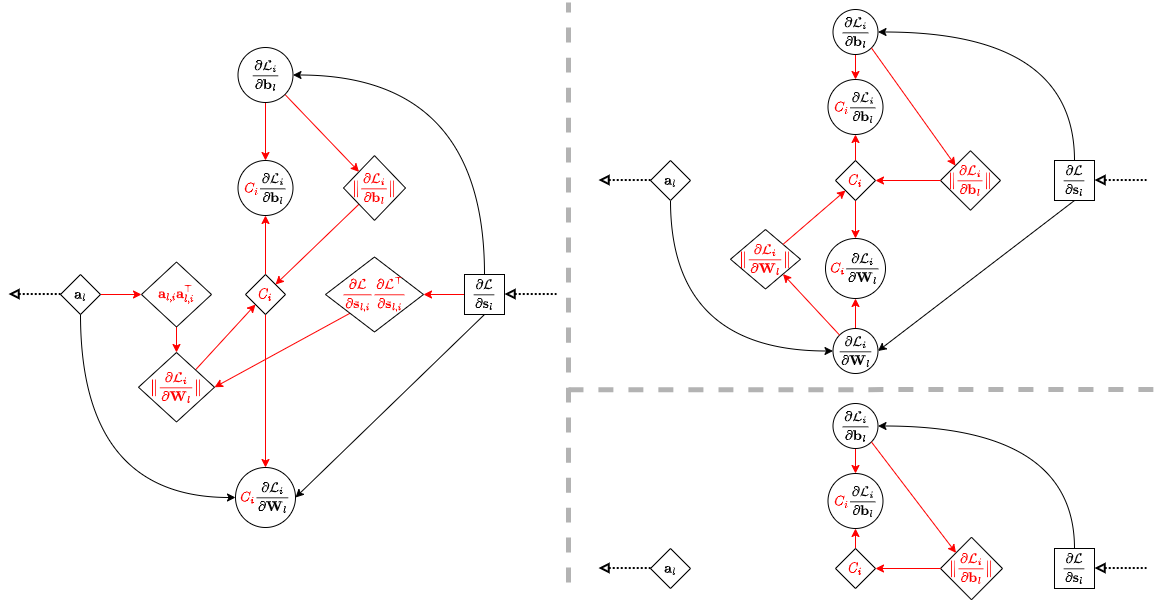}
	\caption{Back-propagation for DP (red\&black) and non-DP (black) algorithms. Note that the bias gradient uses a much simpler computation graph than the weight gradient, rendering DP-BiTFiT easy-to-implement and efficient-to-compute. Left: full fine-tuning with GhostClip (ghost clipping; \citep{goodfellow2015efficient,li2021large,bu2022scalable}). Upper right: full fine-tuning with Opacus \citep{opacus}. Lower right: DP-BiTFiT.}
	\label{fig:flowcharts}
\end{figure*}

We propose DP-BiTFiT, to privately train only the bias terms in a neural network by combining \Cref{eq:outer product linear} and \Cref{eq: private gradient}. We use shaded lines to represent the additional DP operations in \Cref{alg:dp bitfit}, and add DP-related variables and operations in red in the computation graph by \Cref{fig:flowcharts}.

Implementation-wise, DP-BiTFiT is different from all existing DP algorithms (including full, LoRA, Adapter, etc.) that optimize weights, since it does not apply a Pytorch forward hook to store the activation $\a_l$ for all layers. We provide the implementation details of DP-BiTFiT in \Cref{app:implementation}. To give a concrete example, we apply DP-BiTFiT to the RoBERTa-large model on QQP dataset, following the same setting as \citet{li2021large} and using one 40GB A100 GPU. This is the most time-consuming text classification task in our work, taking 119 minutes per epoch for a training batch size 20 using the fastest DP full fine-tuning implementation -- GhostClip \citep{li2021large}. To conduct a simple ablation study, setting all weights to not require gradients (but forward hooks are still operating) reduces the training time by 50\% to to 80 minutes; removing the forward hooks further reduces the training time by 30\% to 63 minutes; finally, using the maximum batch size allowed by the memory-saving DP-BiTFiT reduces to 43 minutes.

\subsection{Parameter efficiency}
DP-BiTFiT enjoys exactly the same parameter efficiency as the standard BiTFiT, training merely about 0.1\% of the total parameters in large models. We demonstrate that DP-BiTFiT is one of the most parameter-efficient fine-tuning through a list of models in \Cref{tab:parameter efficiency}. 

\begin{table}[!htb]
\vspace{-0.4cm}
	\centering
	  \resizebox{0.9\linewidth}{!}{
	\setlength{\tabcolsep}{1pt}
\begin{tabular}{c|c|c|c}
     Dataset&Model&\# of params&\% of bias \\\hline\hline
   \multirow{5}{*}{ImageNet} & VGG16& 138M&0.009\\
& ResNet18& 11.7M&0.043\\
& ResNet50& 25.6M&0.113\\
& ViT-small-patch16& 21.7M&0.238\\
& ViT-base-patch16& 85.8M&0.120\\
& ViT-large-patch16& 303M&0.090\\\hline\hline
\multirow{3}{*}{E2E} & GPT2-small& 124M&0.082\\
&GPT2-medium& 355M&0.076\\
&GPT2-large&774M &0.066\\\hline\hline
\multirow{2}{*}{GLUE} & RoBERTa-base& 125M&0.083\\
&RoBERTa-large &355M &0.077\\
\end{tabular}
}
\vspace{-0.3cm}
\caption{Parameter efficiency of (DP) BiTFiT. Extended results on more models are in \Cref{tab:parameter efficiency extended}.}
\label{tab:parameter efficiency}
\vspace{-0.5cm}
\end{table}

An advantage of this parameter efficiency is reflected in the computation efficiency, given that most parameters do not require gradients to be computed: we show in \Cref{tab:complexity overhead weight bias} and \Cref{sec:scalability of DP} that DP-BiTFiT is much more efficient than full fine-tuning (DP and even non-DP). Additionally, the parameter efficiency also translates to the communication efficiency in the distributed learning. For example, the 64-bit communication cost of DP full fine-tuning is $64MD$ where $M$ is number of worker and $D$ is total number of parameters, which can be reduced $1000\times$ by DP-BiTFiT.

\subsection{Complexity of weight and bias training}

\begin{table*}[!htb]
\centering
\caption{Per-layer time and space complexity (measured by float-point operations) of training on weights (full and LoRA, Adapter; rank$=16$ as in \cite{yu2021differentially}) and biases. Only bias training's overhead is free of $\textcolor{red}{T}$. `$+$' means additional overhead to non-DP training, and `$\langle\rangle$' means between two values.
The layer index $l$ is omitted for simplicity.
}
\vspace{-0.3cm}
\setlength{\tabcolsep}{1pt}
\resizebox{\linewidth}{!}{
    \begin{tabular}{c||c||c|c|c|c|c|c|c|c||c|c}
         &forward&\multicolumn{6}{c|}{weight training}&\multicolumn{2}{|c||}{bias training}\\
         \cline{3-10}
         & \&output grad&non-DP(full)&Opacus(full)&GhostClip(full)
         &Book-Keeping(full)
         &DP(LoRA)&DP(Adapter)
         &non-DP&DP (ours)\\\hline
         \multirow{2}{*}{\shortstack{Time\\complexity}}&\multirow{2}{*}{$4BTpd$}&\multirow{2}{*}{$2BTpd$}&$+2B{\color{red}T}pd$&$+2B{\color{red}T}pd$&\multirow{2}{*}{$O(T)\approx 0$}
&\multirow{2}{*}{$+32B{\color{red}T}(p+d)$}&\multirow{2}{*}{$+64B{\color{red}T}p$}      
&\multirow{2}{*}{$BTp$}&\multirow{2}{*}{$+3Bp$}  \\
         &&&&$ + 2B{\color{red}T^2}(p+d)$&
         &&&&  \\\hline         \multirow{2}{*}{\shortstack{Space\\complexity}}&\multirow{2}{*}{\shortstack{$pd+$\\$BT(p+d)$}}&\multirow{2}{*}{$BT(p+d)$}&\multirow{2}{*}{$+Bpd$}&\multirow{2}{*}{$+2BT^2$}&\multirow{2}{*}{$+\min\{2BT^2,2Bpd\}$}
         &\multirow{2}{*}{$+16B(p+d)$}         &\multirow{2}{*}{$+32Bp$}
         &\multirow{2}{*}{$p$}&\multirow{2}{*}{$+Bp$}
         \\
         &&&&&&&&&
         \\\hline
         \# back-prop&&1&1&2&1
         &1 or 2&1 or 2&1&1
         \\\hline
storing activation&&{\color{red}\cmark}&{\color{red}\cmark}&{\color{red}\cmark}&{\color{red}\cmark}
&{\color{red}\cmark}&{\color{red}\cmark}
&\xmark&\xmark
    \end{tabular}
}
\vspace{-0.4cm}
\label{tab:complexity overhead weight bias}
\end{table*}

We present in \Cref{tab:complexity overhead weight bias} the complexity of DP training on weights and biases, for one layer mapping $B\times T_l\times d_l$ to $B\times T_l\times p_l$. To elaborate on \Cref{footnote3}, for text data, $T_l$ is the sequence length, $d_l$ is input dimension, and $p_l$ is output dimension; for image data and specially in a convolution layer, $T_l$ is height times width, $d_l$ is the input channels times kernel sizes, $p_l$ is the output channels \citep[c.f.][Section 2.3]{bu2022scalable}. Notice that the total complexity of training a network is summed across all layers, e.g. the time complexity of standard full training is $6B\sum_l T_l p_l d_l$, DP full fine-tuning is over $8B\sum_l T_l p_l d_l$, and DP-BiTFiT is about $4B\sum_l T_l p_l d_l$. Therefore, our complexity analysis indicates that DP-BiTFiT is $6/4=1.5\times$ faster than non-private full fine-tuning and over $8/4=2\times$ faster than DP full fine-tuning.

Here, the DP weight training (full fine-tuning or any other PEFT) uses three efficient implementations that are equivalent mathematically but have different complexity: Opacus \citep{opacus}, GhostClip \citep{goodfellow2015efficient,li2021large}, and MixGhostClip \citep{bu2022scalable}. The first two implementations are illustrated in \Cref{fig:flowcharts}, of which MixGhostClip is a hybridization that reduces to GhostClip when $T$ is small. These implementations have been thoroughly analyzed in Appendix C of \cite{bu2022scalable} and we take the complexity result from \citet[Table 1]{bu2022scalable}. For the complexity of bias training in \Cref{tab:complexity overhead weight bias}, it suffices to analyze Line 5 of \Cref{alg:dp bitfit}. We leave the details in \Cref{app:param eff and complexity}, where we also apply the complexity analysis of weight training beyond full fine-tuning, including DP LoRA and DP Adapter for the first time.

\subsection{Scalability of DP algorithms}
\label{sec:scalability of DP}
By \Cref{tab:complexity overhead weight bias}, we observe that DP training on weights can be memory costly, especially when the models are large and the data is high-dimensional. As an example of the large modelling issue, \citet{li2021large} shows that Opacus cannot fit even a single datapoint into a 16GB GPU using GPT2-large \citep{radford2019language} with 774M parameters, due to its $O(B\sum_l p_l d_l)$ space complexity where the number of parameters is $\sum_l p_l d_l$; for high-dimensional data, GhostClip cannot fit a single $400\times 400$ image into the same GPU using ResNet18 with 11.7M parameters, due to its $O(B\sum_l T_l^2)$ space complexity. Although MixGhostClip \citep{bu2022scalable,bu2023differentially} significantly alleviates the memory issue in both cases, the computational overhead from DP training may still be a concern when the dimension is extremely high \citep[c.f.][Figure 4]{bu2022scalable}. In sharp contrast, DP-BiTFiT is amazingly scalable since its computational overhead is negligible and independent of $T$ 
(though the total complexity is still linear in $T$).

\subsubsection{Efficiency v.s. feature dimension}

\begin{figure}[!htb]
\vspace{-0.4cm}
    \centering
    \includegraphics[width=0.48\linewidth]{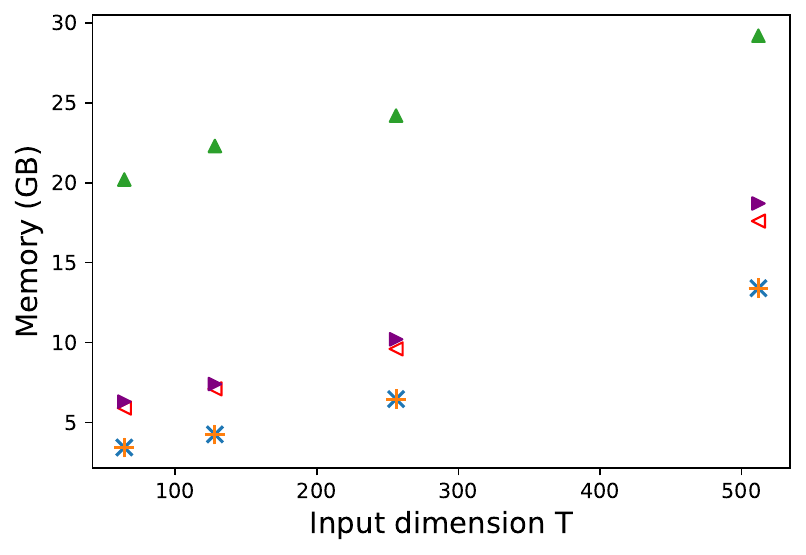}
    \includegraphics[width=0.48\linewidth]{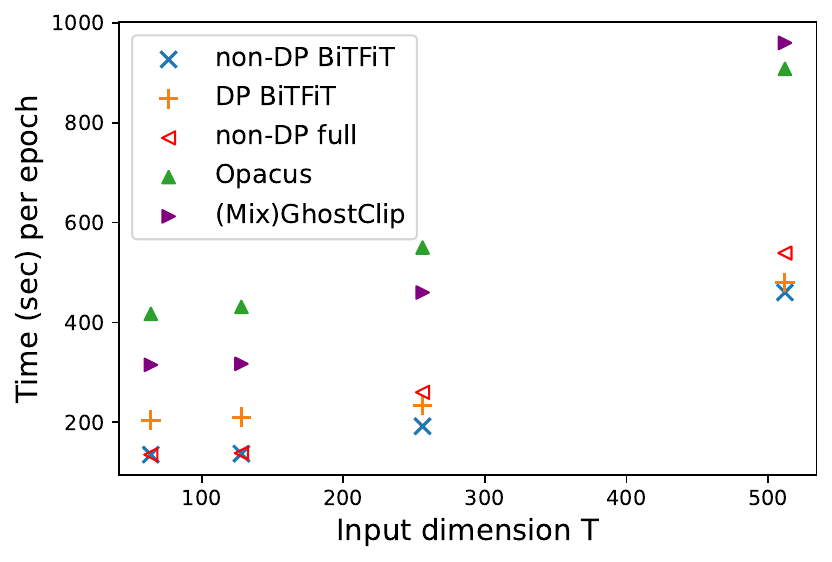}        \includegraphics[width=0.48\linewidth]{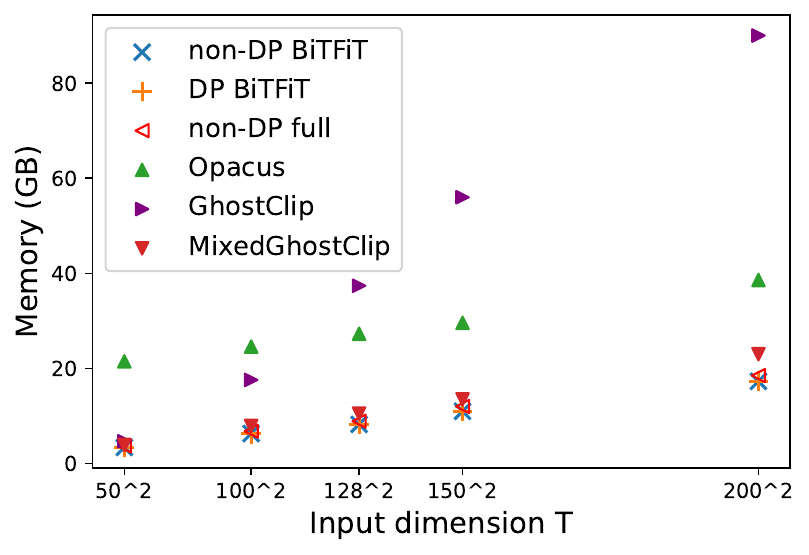}
    \includegraphics[width=0.48\linewidth]{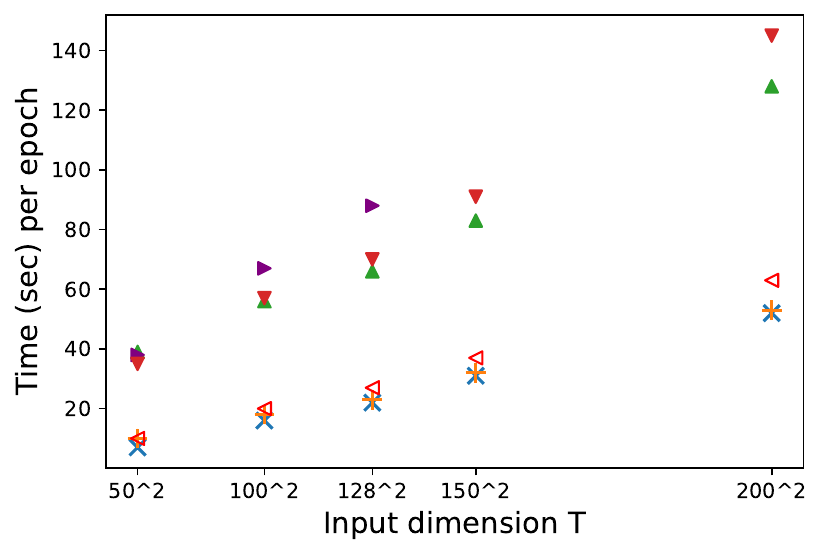} 
    \vspace{-0.3cm}
    \caption{Memory and speed by different fine-tuning methods. Top two: SST2 dataset (sequence length $T$; MixGhostClip is equivalent to GhostClip for this small $T$), RoBERTa-base and batch size 20.
    Bottom two: 50000 images of
    $\sqrt{T}\times\sqrt{T}$ pixels, ResNet50 and batch size 200.}
\label{fig:memory_time_vs_sequence_roberta}
\end{figure}

To empirically evaluate the computation efficiency of DP fine-tuning methods, we measure the time and GPU memory for a fixed batch size. We depict the high-dimensional data issue in \Cref{fig:memory_time_vs_sequence_roberta}, in which the memory saving and speedup by DP-BiTFiT is substantial. We expect to observe greater efficiency advantage of DP-BiTFiT on higher dimensional data, e.g. in document-level language tasks with $T\approx 20000$ by \citet{beltagy2020longformer}, and in high-resolution image tasks, such as $1024\times 1024$ CelebA-HQ \citep{karras2018progressive} and Flickr-Faces-HQ \citep{karras2019style} where $T$ can be of order $10^5$ in the convolution layers.

\begin{table*}[!b]
\vspace{-0.2cm}
\centering
\caption{Accuracy of fine-tuning methods with RoBERTa, under $\epsilon=8$. More non-private fine-tuning results (similar to here) can be found in \citep{yu2021differentially,hu2021lora,zaken2022bitfit}. Note that linear probing of RoBERTa-base only gets 87.2\% on SST2 and 77.3\% on QNLI.}	
    \vspace{-0.3cm}
\setlength{\tabcolsep}{0.8pt}
	\resizebox{0.9\linewidth}{!}{
		\begin{tabular}{|c||c|c|c|c|c|c|c|c|c|}			
	    \hline
		& \multicolumn{2}{|c|}{Full }&RGP&Adapter&\multicolumn{2}{|c|}{LoRA}&\multicolumn{2}{|c|}{BiTFiT}&Compacter \\
		& \multicolumn{2}{|c|}{\citep{li2021large}}&\citep{yu2021differentially}&\citep{yu2021differentially}&\multicolumn{2}{|c|}{ \citep{yu2021differentially}}&\multicolumn{2}{|c|}{Ours}& \citep{yu2021differentially} \\\hline
			Additional params to networks&\multicolumn{2}{|c|}{\xmark}&\xmark&\cmark&\multicolumn{2}{|c|}{\cmark}&\multicolumn{2}{|c|}{\xmark}&\cmark 
			\\\hline
			Forward caching activations&\multicolumn{2}{|c|}{\cmark}&\cmark&\cmark&\multicolumn{2}{|c|}{\cmark}&\multicolumn{2}{|c|}{\xmark}&\cmark
			\\\hline\hline
			\multicolumn{10}{|c|}{RoBERTa-base (125M)}
			\\\hline
			\% of trainable params&\multicolumn{2}{|c|}{100\%}&100\%&1.4\%&\multicolumn{2}{|c|}{0.94\%}&\multicolumn{2}{|c|}{0.083\%}&0.055\%
			\\\hline
			&\textcolor{cyan}{standard}&DP&DP&DP&\textcolor{cyan}{standard}&DP&\textcolor{cyan}{standard}&DP&DP\\\hline
			Accuracy SST2&\textcolor{cyan}{94.5}&92.1&91.6&92.5&\textcolor{cyan}{95.1}&92.2&\textcolor{cyan}{93.5}&92.4&92.3
			\\
			Accuracy QNLI&\textcolor{cyan}{91.4}&87.9&87.2&87.5&\textcolor{cyan}{93.3}&87.3&\textcolor{cyan}{87.3}&86.9&85.1
			\\
			Accuracy QQP&\textcolor{cyan}{87.3}&86.1&85.5&85.6&\textcolor{cyan}{90.8}&85.7&\textcolor{cyan}{86.1}&85.6&84.7
			\\
			Accuracy MNLI-m&\textcolor{cyan}{85.9}&83.2&80.1&83.4&\textcolor{cyan}{87.5}&83.5&\textcolor{cyan}{83.4}&82.9&82.6
			\\\hline\hline
		\multicolumn{10}{|c|}{RoBERTa-large (355M)}
			\\\hline
			\% of trainable params&\multicolumn{2}{|c|}{100\%}&100\%&1.4\%&\multicolumn{2}{|c|}{0.94\%}&\multicolumn{2}{|c|}{0.077\%}&0.053\%
			\\\hline
			&\textcolor{cyan}{standard}&DP&DP&DP&\textcolor{cyan}{standard}&DP&\textcolor{cyan}{standard}&DP&DP\\\hline
			Accuracy SST2&\textcolor{cyan}{96.2}&93.8&93.0&93.9&\textcolor{cyan}{96.2}&95.3&\textcolor{cyan}{95.5}&94.5&94.2
			\\
			Accuracy QNLI&\textcolor{cyan}{93.6}&91.1&90.0&90.7&\textcolor{cyan}{94.9}&90.8&\textcolor{cyan}{92.2}&91.1&90.2
			\\
			Accuracy QQP&\textcolor{cyan}{87.9}&87.5&86.7&86.3&\textcolor{cyan}{91.6}&87.4&\textcolor{cyan}{87.9}&86.9&86.2
			\\
			Accuracy MNLI-m&\textcolor{cyan}{90.3}&87.0&86.1&87.7&\textcolor{cyan}{90.6}&87.8&\textcolor{cyan}{89.3}&88.3&87.5
			\\\hline
		\end{tabular}
	}
\label{tab:finetuning comparison}
\end{table*}

\subsubsection{Efficiency v.s. model size}
\begin{figure}[!htb]
\vspace{-0.7cm}
    \centering
    \includegraphics[width=0.48\linewidth]{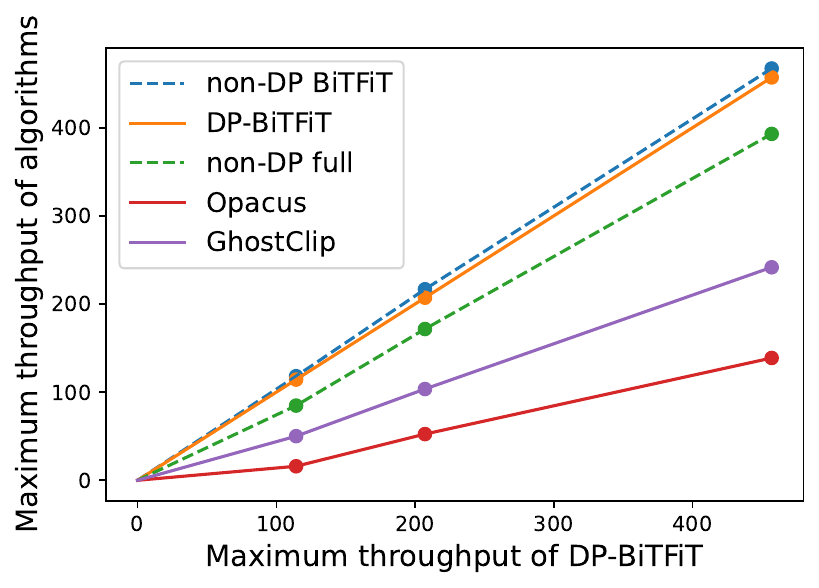}
\includegraphics[width=0.48\linewidth]{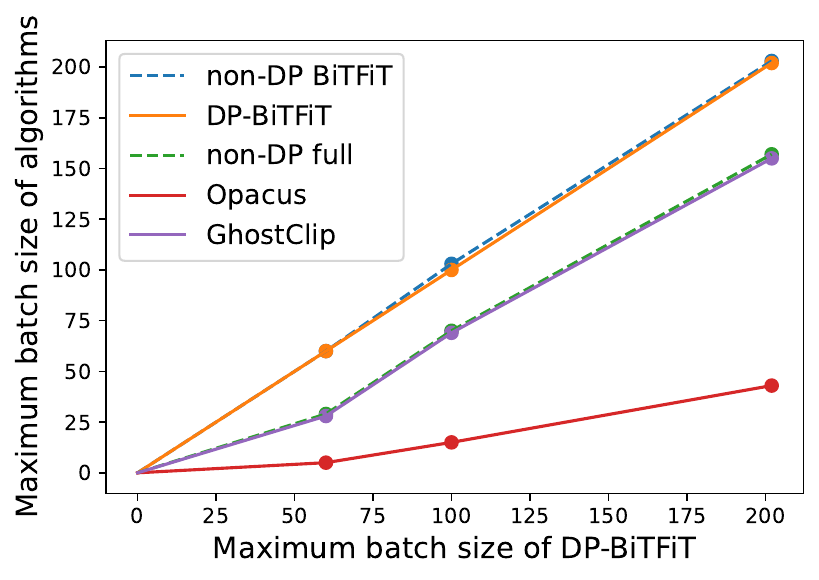}
    \includegraphics[width=0.48\linewidth]{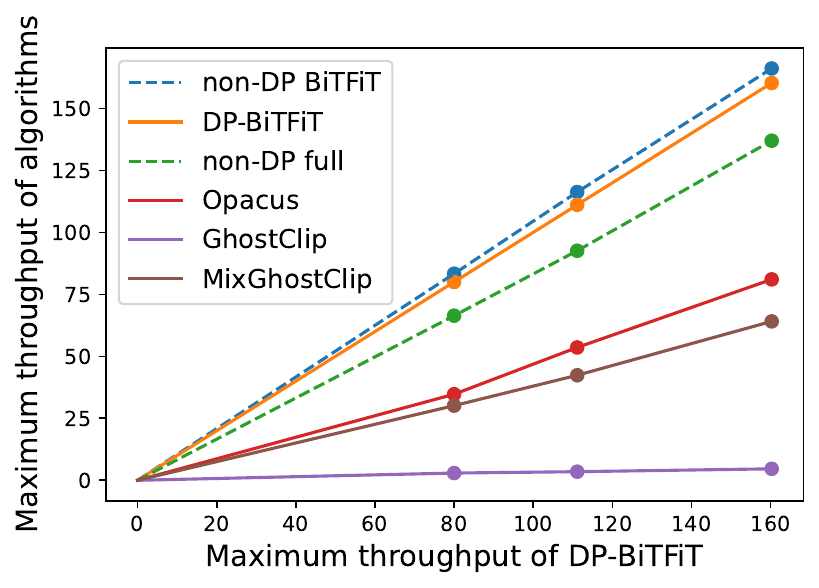}
    \includegraphics[width=0.48\linewidth]{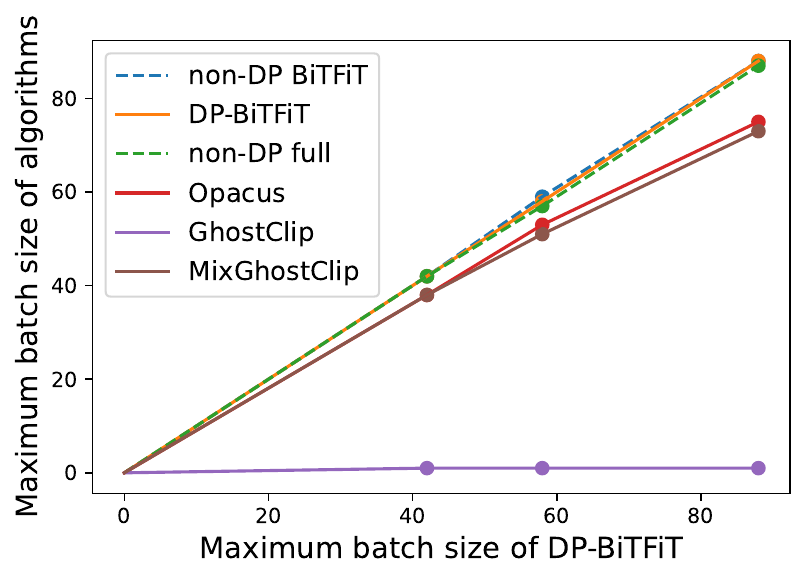}
    \vspace{-0.3cm}
\caption{Maximum throughput and batch size by different fine-tuning methods. Each model is represented by one column, which sorts the model size in decreasing order from left to right.
Top two: E2E dataset with GPT2-small/medium/large (MixGhostClip is equivalent to GhostClip for this small $T$). Bottom two: 50000 images of $512\times 512$ pixels with ResNet 50/101/152. 
}
\label{fig:memory_time_vs_models}
\end{figure}

To stress-test the computation efficiency of DP-BiTFiT with large models, we apply the maximum batch size with respect to each fine-tuning method, instead of using a fixed one across different methods. Therefore, DP-BiTFiT can further leverage its memory efficiency to achieve the best throughput. Here we consider a setting of high-dimensional data ($T=512^2$) but small ResNet ($11.7\sim 58.2$M parameters) and the other setting of low-dimensional data ($T=100$) but large GPT2 ($125\sim 774$M parameters).

\subsection{Applicability of DP-BiTFiT}

Some model architectures, such as LLAMA \cite{touvron2022llama,touvron2023llama,chowdhery2023palm} and the convolutional layers followed by batch normalization \cite{he2016deep}, may not contain any bias terms, hence DP-BiTFiT (and its non-DP counter-part) is either not directly applicable or less performant. We propose DP-BiTFiT-Add, which adds zero or randomly initialized bias terms to the layers and then applies DP-BiTFiT. Such initialization does not affect the pre-trained utility, but enlarges the parameter space and thus allows better performance after fine-tuning. As a concrete example, we experiment with ResNet18 (no bias in all convolutional layers) on CelebA for the multi-label classification task, under the same setting as \Cref{tab:finetuning images}. The accuracy boosts from 86.9\% of DP-BiTFiT to 87.3\% of DP-BiTFiT-Add, compared to 88.4\% of the full fine-tuning. 

Note that DP-BiTFiT-Add is still activation-free and highly parameter-efficient: DP-BiTFiT-Add trains less than 0.1\% parameters on ResNet18, and only 0.03\% parameters on LLAMA2-7B.

\section{Experiments}
\label{sec:experiments}
We now test the accuracy of DP-BiTFiT on natural language and computer vision tasks, with the settings in \Cref{app:experiment details}. For DP full fine-tuning algorithms, we use GhostClip \citep{li2021large} on texts, and MixedGhostClip \citep{bu2022scalable} on images, which achieve SOTA efficiency and accuracy on these datasets respectively. We compute $\epsilon$ using a conversion from RDP though tighter privacy accountants in \Cref{sec:dp prelim} are feasible. 
And we observe that, in all experiments with or without DP, the optimal learning rate for BiTFiT is larger than that for full fine-tuning.

\subsection{Text classification}
We experiment on MNLI-m(mismatch) \citep{N18-1101}, QQP \citep{WinNT}, QNLI \citep{rajpurkar2016squad}, and SST2 datasets \citep{socher2013recursive}. Competitive algorithms include reparameterized gradient perturbation (RGP, \citep{yu2021large}), LoRA, Adapter and Compacter \citep{yu2021differentially}. We use the same setup as \citet{li2021large} on RoBERTa models with text-infiling, only increasing the learning rate for DP-BiTFiT. Additional results under a stronger privacy guarantee $\epsilon=3$ can be found in \Cref{tab:finetuning comparison eps3}.

In \Cref{tab:finetuning comparison}, DP-BiTFiT is highly parameter efficiency and accurate compared with other DP fine-tuning. As indicated by \Cref{fig:all param efficient} and  \Cref{fig:memory_time_vs_sequence_roberta}, over $2\times$ speedup and over $3\times$ memory saving is observed, when switching from DP full fine-tuning to DP-BiTFiT.

\begin{remark}\label{rem:gap better large model}
It is encouraging to observe that the gap between the full fine-tuning and BiTFiT, with or without DP, tends to decrease as the model size increases. For instance on QNLI, this gap without privacy reduces from 4.1\% to 1.4\%, and with privacy reduces from 1.4\% to 0.1\%. This scaling pattern is consistently observed on different tasks, e.g. in \Cref{tab:gpt e2e} and \Cref{tab:finetuning images}.
\end{remark}

\subsection{Natural Language Generation}

\begin{table*}[!htb]
	\centering
\caption{Performance of fine-tuning methods with GPT2, under $\epsilon=8$. LoRA and prefix results are documented in \citet{li2021large}. Best performance in each model is in bold text. DP-BiTFiT is comparable to DP full, especially on larger models.}
\vspace{-0.3cm}
\setlength{\tabcolsep}{2pt}
	\resizebox{0.86\linewidth}{!}{
		\begin{tabular}{c|c|c|c|c|c|c|c|c|c}
			Model&Fine-tuning&\% of params&Privacy&Perplexity$\downarrow$&BLEU$\uparrow$&ROGUE-L$\uparrow$&NIST$\uparrow$&METEOR$\uparrow$&CIDEr$\uparrow$\\\hline
			\multirow{8}{*}{\shortstack{GPT2-small\\(124M)}}&\multirow{2}{*}{full}&\multirow{2}{*}{100\%}&\textcolor{cyan}{standard}&\textcolor{cyan}{2.91}&\textcolor{cyan}{69.46} &\textcolor{cyan}{71.36} &\textcolor{cyan}{8.78} &\textcolor{cyan}{0.46}&\textcolor{cyan}{2.42}\\
			&&&DP ($\epsilon=8$)&2.33&\textbf{63.60}&67.07&\textbf{7.71}&0.40&1.94\\\cline{2-10}
			&\multirow{2}{*}{LoRA}&\multirow{2}{*}{---}&\textcolor{cyan}{standard}&\textcolor{cyan}{---}&\textcolor{cyan}{69.68} &\textcolor{cyan}{71.71} &\textcolor{cyan}{8.82} &\textcolor{cyan}{0.46}&\textcolor{cyan}{2.49}\\
			&&&DP ($\epsilon=8$)&---&63.39&\textbf{67.53}&7.45&\textbf{0.41}&\textbf{1.95}\\\cline{2-10}
			&\multirow{2}{*}{prefix}&\multirow{2}{*}{---}&\textcolor{cyan}{standard}&\textcolor{cyan}{---}&\textcolor{cyan}{68.85} &\textcolor{cyan}{70.81} &\textcolor{cyan}{8.72} &\textcolor{cyan}{0.45}&\textcolor{cyan}{2.35}\\
			&&&DP ($\epsilon=8$)&---&49.26&60.73&5.53&0.36&1.57\\\cline{2-10}
			&\multirow{2}{*}{BiTFiT}&\multirow{2}{*}{0.082\%}&\textcolor{cyan}{standard}&\textcolor{cyan}{3.19}&\textcolor{cyan}{64.46} &\textcolor{cyan}{63.67} &\textcolor{cyan}{4.25} &\textcolor{cyan}{0.36}&\textcolor{cyan}{1.36}\\
			&&&DP ($\epsilon=8$)&2.89&60.56&64.96&6.14&0.37&1.62\\\hline\hline
			\multirow{4}{*}{\shortstack{GPT2-medium\\(355M)}}&\multirow{2}{*}{full}&\multirow{2}{*}{100\%}&\textcolor{cyan}{standard}&\textcolor{cyan}{2.08}&\textcolor{cyan}{68.50} &\textcolor{cyan}{71.46} &\textcolor{cyan}{8.63} &\textcolor{cyan}{0.45}&\textcolor{cyan}{2.14}\\
			&&&DP ($\epsilon=8$)&2.25&\textbf{64.22}&\textbf{67.53}&\textbf{8.17}&\textbf{0.42}&\textbf{2.08}\\\cline{2-10}
			&\multirow{2}{*}{BiTFiT}&\multirow{2}{*}{0.076\%}&\textcolor{cyan}{standard}&\textcolor{cyan}{2.85}&\textcolor{cyan}{64.48} &\textcolor{cyan}{67.81} &\textcolor{cyan}{8.50} &\textcolor{cyan}{0.43}&\textcolor{cyan}{2.11}\\
			&&&DP ($\epsilon=8$)&2.67&61.02&66.13&7.18&0.39&1.80\\\hline\hline
			\multirow{4}{*}{\shortstack{GPT2-large\\(774M)}}&\multirow{2}{*}{full}&\multirow{2}{*}{100\%}&\textcolor{cyan}{standard}&\textcolor{cyan}{1.79}&\textcolor{cyan}{66.84} &\textcolor{cyan}{70.38} &\textcolor{cyan}{8.73} &\textcolor{cyan}{0.46}&\textcolor{cyan}{2.36}\\
			&&&DP ($\epsilon=8$)&2.26&64.64&\textbf{68.97}&8.30&\textbf{0.42}&\textbf{2.16}\\\cline{2-10}
			&\multirow{2}{*}{BiTFiT}&\multirow{2}{*}{0.066\%}&\textcolor{cyan}{standard}&\textcolor{cyan}{2.79}&\textcolor{cyan}{65.79} &\textcolor{cyan}{67.61} &\textcolor{cyan}{8.55} &\textcolor{cyan}{0.43}&\textcolor{cyan}{2.21}\\
	&&&DP ($\epsilon=8$)&2.59&\textbf{65.21}&67.88&\textbf{8.43}&\textbf{0.42}&2.15\\\hline
	\end{tabular}
	}
	\label{tab:gpt e2e}
\end{table*}

We compare DP-BiTFiT with DP LoRA, full fine-tuning, and prefix tuning \citep{li2021prefix} on E2E dataset \citep{dusek.etal2020:csl}, in order to train GPT2 that generates texts to evaluate a restaurant. The performance measures are BLEU \citep{Papineni02bleu}, ROGUE-L \citep{lin-2004-rouge}, NIST \citep{sadjadi20182017}, METEOR \citep{banarjee2005}, CIDEr \citep{vedantam2015cider} and perplexity. We use the same setup as \citet{bu2022automatic} with automatic clipping, only increasing the learning rate for DP-BiTFiT. More results under a stronger privacy guarantee $\epsilon = 3$ can be found in \Cref{tab:gpt e2e eps3}.

In \Cref{tab:gpt e2e}, DP-BiTFiT has shown strong performance, even outperforming DP full fine-tuning on GPT2-large, as well as both the computation and parameter efficiency (see \Cref{fig:memory_time_vs_models}). Similar to \Cref{rem:gap better large model}, the gap of BLEU score between DP-BiTFiT and DP full fine-tuning reduces from -3.06/-3.20 (GPT2-small/medium) to +0.57 (GPT2-large), as the model size increases. We refer to \Cref{tab:gpt e2e eps3} for a more significant pattern when $\epsilon=3$.

\begin{table}[!htb]
\centering
\setlength{\tabcolsep}{1pt}
\caption{Accuracy of DP ViT-large on CIFAR, 3 epochs.}
\vspace{-0.3cm}
\resizebox{0.72\linewidth}{!}{
\begin{tabular}{|c|c|c|c|}
\hline
CIFAR10&DP last-layer&DP-BiTFiT&DP full
\\\hline
$\epsilon=1$&98.4&  98.9& 98.9
\\
$\epsilon=2$&98.6&  99.0& 98.9
\\
$\epsilon=4$&98.6&  99.0& 99.0
\\
$\epsilon=8$&98.7&99.0& 99.0
\\\hline
CIFAR100&DP last-layer&DP-BiTFiT&DP full
\\\hline
$\epsilon=1$&86.2&90.2&87.7
\\
$\epsilon=2$&87.3&91.2&90.1
\\
$\epsilon=4$&88.1&91.8&91.0
\\
$\epsilon=8$&88.8&92.3&91.3
\\\hline
\end{tabular}
}
\label{tab:cifar100 BEIT}
\\
\hspace{-0.9cm}
\includegraphics[width=0.7\linewidth]{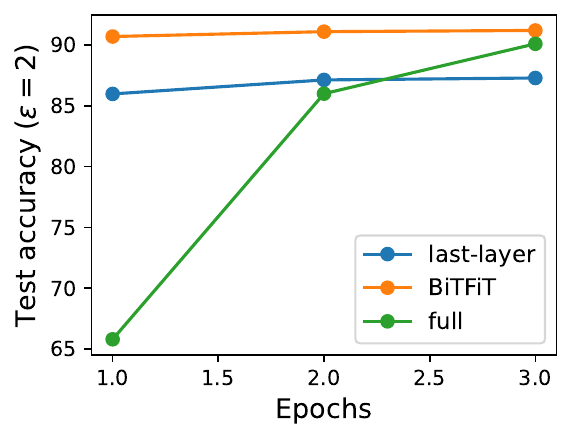}
\vspace{-0.4cm}
\captionof{figure}{Accuracy of DP ViT-large on CIFAR100.}
\end{table}

\subsection{Image classification}
\label{sec:image class}

We further experiment on CIFAR10/CIFAR100 ($32\times 32$ pixels, resized to $224\times 224$) and CelebA ($218\times 178$ pixels, not resized; results in \Cref{tab:celebA 40} and \Cref{tab:finetuning images}) after pre-training on ImageNet ($224\times 224$ pixels). For these downstream datasets (e.g. CIFAR10 has only 10 classes), the number of classes is different than that in ImageNet, which has 1000 classes. Consequently, the classification head of the pretrained model is re-placed by random initialization. Therefore, our DP-BiTFiT is applied on top of the last-layer training, but the number of trainable parameter remains $\approx 0.1\%$ of the model parameters. For instance, ViT-large has 303M parameters, of which 282k are biases and the weight of last layer contains $\approx 100$k, depending on the number of classes in the downstram task.

We observe that DP-BiTFiT enjoys $1.5\times$ speedup for transformers and ResNet in \Cref{tab:celebA 40}, and that DP-BiTFiT performs on par with full fine-tuning in \Cref{tab:finetuning images,tab:cifar100 BEIT,tab:extended two-phase-cifar10,tab:extended two-phase-cifar100}, e.g. achieving state-of-the-art 99.0\% accuracy on CIFAR10 and 91.2\% on CIFAR100 at $\epsilon=2$. Our observation holds across various models (especially on transformers), privacy budgets, and datasets. However, DP-BiTFiT needs extra attention for convolutional neural networks (CNN) as we elaborate in \Cref{app:bias cnn}.

\begin{table*}[!htb]
\centering
\setlength{\tabcolsep}{1.5pt}
\caption{Accuracy of DP fine-tuning methods on CIFAR10 and CelebA. More results under different $\epsilon$ and network architectures can be found in \Cref{app:more results image}.}
		\begin{tabular}{|c||c|c|c|c|}
	    \hline
	    Dataset&&Model&Fine-tuning&Accuracy
	    \\\hline
		\multirow{7}{*}{\shortstack[c]{CIFAR10\\ ($\epsilon=2,\delta=$1e-5)}}
		&\citep{yu2021do}&ResNet152 (GEP)&last-layer&94.8\\\cline{2-5}
		&\citep{tramer2020differentially}&SIMCLRv2&last-layer&92.7\\\cline{2-5}
		&\multirow{2}{*}{\citep{de2022unlocking}}&Wide-ResNet28&last-layer&93.6\\
		&&Wide-ResNet28&full&95.4\\\cline{2-5}
		&\multirow{3}{*}{\citep{bu2022scalable}}&crossvit-base-240&full&96.1\\
  &&vit-base-patch16&full&97.4\\
  \cline{2-5}
		&&vit-large-patch16&full&98.9\\
		&\multirow{3}{*}{Ours}&crossvit-base-240&BiTFiT&95.7
		\\
		&&vit-base-patch16&BiTFiT& 97.7
		\\
		&&vit-large-patch16&BiTFiT& 99.0
		\\\hline
		\multirow{3}{*}{\shortstack[c]{CelebA [Smiling]\\($\epsilon=8,\delta=$5e-6)}}&\citep{bu2022automatic}&ResNet9&full&91.08\\\cline{2-5}
		&\multirow{3}{*}{Ours}&ResNet18&full&91.02\\
		&&ResNet18&BiTFiT&88.17\\
		&&ResNet18&last-layer&66.15\\\hline
		\multirow{3}{*}{\shortstack[c]{CelebA [Male]\\ ($\epsilon=8,\delta=$5e-6)}}&\citep{bu2022automatic}&ResNet9&full&95.70\\\cline{2-5}
		&\multirow{3}{*}{Ours}&ResNet18&full&95.15\\
		&&ResNet18&BiTFiT&92.29\\
		&&ResNet18&last-layer&78.70\\
		\hline
		\multirow{3}{*}{\shortstack[c]{CelebA [Multi-label]\\ ($\epsilon=8,\delta=$5e-6)}}&\citep{bu2022automatic}&ResNet9&full&87.58\\\cline{2-5}
		&\multirow{3}{*}{Ours}&ResNet18&full&88.38\\
		&&ResNet18&BiTFiT&86.87\\
		&&ResNet18&last-layer&83.67\\\hline
		\end{tabular}
	\label{tab:finetuning images}
\end{table*}

\vspace{-0.4cm}

\section{Discussion}
In this work, we study DP-BiTFiT to privately train the bias terms of neural networks. The highlight of DP-BiTFiT is the accuracy, the parameter efficiency and the computation efficiency, which is realized by not forward caching the activation tensors, and not back-propagating the gradient of weights. This unique mechanism allows DP-BiTFiT to be as fast and memory-saving as its non-private counterpart, thus particularly suitable for large models and/or high-dimension data on which the full fine-tuning can be costly.

While we have studied DP-BiTFiT as a standalone method, it is promising to combine it with other methods, such as prefix-based tuning and \textit{weights}-based fine-tuning. For instance, one can fine-tune DP LoRA+BiTFiT, via $f(\x;\W_0,\hat\b,\hat{\bm\theta})$ to obtain even better performance\footnote{In fact, this has been acknowledged in the non-DP LoRA \cite{hu2021lora}: "\href{https://github.com/microsoft/LoRA}{Training bias vectors in tandem with LoRA might be a cost-efficient way to squeeze out extra task performance}".}. We readily offer such flexible combination in our codebase, which automatically implements any DP algorithms in the backend.

\clearpage
\section*{Impact Statement}
This paper presents work whose goal is to advance the field of Machine Learning. There are many potential societal consequences of our work, none which we feel must be specifically highlighted here.

\bibliography{reference}
\bibliographystyle{icml2024}

\newpage
\appendix
\onecolumn
\clearpage

\section{Detailed analysis}
\subsection{Back-propagation}
\label{app:detail back-propagation}
We rigorously analyze the neural network represented in \Cref{sec:backward}: for sample index $i\in [B]$,
\begin{align}
	\underbrace{\a_{l+1,i}}_{\R^{T\times d'}} = \phi (\underbrace{\s_{l,i}}_{\R^{T\times p}}),\quad\quad  \s_{l,i}=\underbrace{\a_{l,i}}_{\R^{T\times d}}\underbrace{\W_{l}}_{\R^{d\times p}}+{\underbrace{\mathbf{1}}_{\R^{T\times 1}}}\cdot\underbrace{\b_{l}}_{\R^{1\times p}},
	\label{eq:linear forward detailed}
\end{align}

Then the per-sample weight gradient is given by the chain rule as
$$	\frac{\partial \mathcal{L}_i}{\partial \W_{l}}^\top=\sum_j\frac{\partial \mathcal{L}_i}{\partial \s_{l,j}}^\top\frac{\partial \s_{l,j}}{\partial \W_{l}}=\frac{\partial \mathcal{L}_i}{\partial \s_{l,i}}^\top\frac{\partial \s_{l,i}}{\partial \W_{l}}=\frac{\partial \mathcal{L}_i}{\partial \s_{l,i}}^\top\a_{l,i}=\frac{\partial \mathcal{L}}{\partial \s_{l,i}}^\top\a_{l,i}
$$
in which the second equality holds when there is no parameter sharing (so that each per-sample loss only depends on $i$-th input and output). The last equality holds for the same reason.

Similarly, we have the per-sample bias gradient as 
$$	\frac{\partial \mathcal{L}_i}{\partial \b_{l}}^\top=\sum_j\frac{\partial \mathcal{L}_i}{\partial \s_{l,j}}^\top\frac{\partial \s_{l,j}}{\partial \b_{l}}=\frac{\partial \mathcal{L}_i}{\partial \s_{l,i}}^\top\frac{\partial \s_{l,i}}{\partial \b_{l}}=\frac{\partial \mathcal{L}_i}{\partial \s_{l,i}}^\top\mathbf{1}=\frac{\partial \mathcal{L}}{\partial \s_{l,i}}^\top\mathbf{1}.
$$

We additionally demonstrate that bias gradient is independent of the input $\a_l$, on the convolution (1d/2d/3d) and the normalization layers. For the convolution, $\s_l$ is the inversely folded output and $\a_l$ is the unfolded input, then the forward pass is the same as that of linear layer in \Cref{eq:linear forward detailed}. Notice that $T$ is the product of hidden feature dimension (c.f. \cite{bu2022scalable}), which depends on the padding, kernel sizes, strides, etc. For the batch, layer, group, and instance normalization, the forward pass is
$$\s_{l,i}=\frac{\a_{l,i}-\mathbb{E}(\a_l)}{\sqrt{\text{Var}(\a_l)+0.00001}}\cdot \W_l+\mathbf{1}\cdot\b_l$$
which can be analyzed similarly to that of \Cref{eq:linear forward detailed}.

\subsection{Making BiTFiT work with convolutional neural networks}
\label{app:bias cnn}
Most (non-transformer) vision models use convolution layers and batch normalization during their standard non-DP training, which is problematic for DP training in general, especially for DP-BiTFiT. We take ResNet \citep{he2016deep} as a concrete example.

Firstly, it is well-known that DP training does not support batch normalization, because the mean and standard deviation are computed based on samples (c.f. \url{https://opacus.ai/tutorials/guide_to_module_validator}). Therefore, in DP training, ResNet-BN (with batch normalization) is modified to a different achitecture ResNet-GN (replaced by group normalization, e.g. \cite{abadi2016deep}). Put differently, ResNet is different in DP and non-DP training and sometimes the comparison may be unfair. This makes vision transformers favorable because they use layer normalization so that the architecures do not require modification when switching to DP regime.

Secondly, the convolution layers usually do not contain bias terms when followed by batch normalization. This is the case in packages like tensorflow.keras, torchvision, timm, and in models like ResNet, ResNext, DenseNet, etc. The reason of not having bias terms is that the batch normalization performs mean subtraction, which make the biases ineffective (see \url{https://discuss.pytorch.org/t/no-bias-in-the-pretrianed-state-dictionary-of-resnet18/153263/2}). In words, ResNet-BN(with bias)$=$ResNet-BN(no bias), but ResNet-GN(with bias)$\neq$ResNet-GN(no bias).

\paragraph{Consequences}
Consider two networks, ResNet(no bias) with bias-less convolution and ResNet(with bias). In full fine-tuning, we are training all 100 layers of both ResNets and they are equivalent under batch normalization; but in DP-BiTFiT, we are essentially not training ResNet(no bias), maybe except the classification head. 

\subsubsection{Walk-around 1}
To walk around, we can manually re-write the convolution layers in CNNs, which is technically troublesome and has to be done in a case-by-case manner. For example, in \cite{bu2022automatic}, ResNet9 was implemented with bias in the convolution layers. This walk-around can improve the performance of DP-BiTFiT significantly (because all layers are trainable now) without sacrificing the training efficiency.

\subsubsection{Walk-around 2}
Alternatively, we can leverage a two-phase training to interpolate between full fine-tuning and BiTFiT. We introduce the \textit{two-phase training}, denoted as $X+$BiTFiT, which firstly applies DP full fine-tuning for $X$ epochs then DP-BiTFiT for the rest of training. Hence, $X$+BiTFiT becomes DP full fine-tuning when $X$ equals total epochs, and reduces to DP-BiTFiT when $X=0$. Empirically speaking, it suffices to use $X\leq 2$ to achieve comparable accuracy to full fine-tuning, while still enjoying some speedup. The effectiveness of two-phase training is verified in \Cref{app:more results image}. 1+BiTFiT outperforms previous SOTA by DP full fine-tuning \cite{bu2022scalable} that used BEiT-large: CIFAR10 $97.1\%\to 98.8\%$; CIFAR100 $86.2\%\to 88.7\%$, under $\epsilon=2$. 2+BiTFiT is comparable to previous SOTA, $87.05/87.58\%\to 86.54/86.71\%$ on CelebA in \Cref{tab:celebA 40}, under $\epsilon=3/8$ respectively.

As a concrete example, our experiments on CIFAR10 shows that while training ViT-tiny with DP-BiTFiT only achieves 82.6\% accuracy, the two-phase training that applies DP full fine-tuning for a single epoch boosts the accuracy to 92.6\%. This boost is even more effective on CIFAR100, where DP-BiTFiT achieves 12\% accuracy but the two-phase training gives 63\%. A number of experiments can be found in \Cref{app:more results image}.

\section{Implementation of DP-BiTFiT}
\label{app:implementation}
In this section we describe the implementation of DP-BiTFiT, which only uses Pytorch backward hook but not the forward hook, and thus is different from existing packages such as FastGradClip \cite{lee2020scaling}, Opacus \cite{opacus}, Private Transformers \cite{li2021large}, Private CNN \cite{bu2022scalable}. Notice that in these packages, the forward hook is used to store the activation tensor $\a_l$ for all layers, which incurs huge memory burden as discussed in \Cref{sec:forward}.

The \href{https://pytorch.org/tutorials/beginner/former_torchies/nnft_tutorial.html}{Pytorch backward hook} is a function, to be registered on a torch Module (or a layer in the neural network), that will be executed in the backward propagation. The backward hook automatically extracts the input gradient $\frac{\partial\mathcal{L}}{\partial\a_l}$ and the output gradient $\frac{\partial\mathcal{L}}{\partial\s_l}$ of the layer. 

In DP-BiTFiT, we call \texttt{register\_backward\_hook} to register a backward hook for Line 5 of \Cref{alg:dp bitfit}. An example for a linear layer: $\R^{B\times T\times d}\to\R^{B\times T\times p}$ looks like

{\scriptsize
\begin{verbatim}
def hook(linear_layer, grad_input, grad_output):
    linear_layer.bias.grad_sample = grad_output.sum(dim=1)
    linear_layer.bias.norm_sample = linear_layer.bias.grad_sample.norm(2,dim=1)
\end{verbatim}
}

Here the attribute \texttt{norm\_sample} stores the per-sample gradient norm $\left\|\frac{\partial \mathcal{L}_i}{\partial \b_l}\right\|_F$, and the attribute \texttt{grad\_sample} stores the $\R^{B\times p}$ per-sample gradient of bias.

Then the implementation of DP-BiTFiT for one iteration looks like
{\scriptsize
\begin{verbatim}
output=model(input)
loss=F.cross_entropy()(output,label)
torch.autograd.grad(loss,biases)
all_layer_norm_sample = torch.stack([param.norm_sample for param in biases],dim=0).norm(2, dim=0)
clipping_factor=1/(all_layer_norm_sample+0.01)
for layer in model.modules():
    layer.bias.grad=torch.einsum("i,i...->...", clipping_factor,layer.bias.grad_sample)
optimizer.step()
optimizer.zero_grad()
\end{verbatim}
}
where \texttt{biases} is the collection of all bias terms in all layers.

\section{Complexity analysis}
\label{app:param eff and complexity}
We provide more details on analyzing the time and space complexity. The analysis for full fine-tuning has been presented in Appendix C of \cite{bu2022scalable}. At high level, the major components of time complexity is from the matrix/tensor multiplication, for example, if a layer takes in $\a_l\in\R^{B\times T\times d}$ and multiply with its $\W_l\in\R^{d\times p}$, the time complexity would be $2BTdp$ for this forward pass, and the back-propagation roughly takes 2 times the time complexity, leading to $(2+4)=6BTdp$ complexity. In some DP algorithms, like GhostClip, the back-propagation is done twice, hence it's roughly $(2+4+4)=10BTdp$.

This analysis is adapted here for the parameter efficient fine-tuning: for example, Adapter \cite{houlsby2019parameter} uses two matrices $W_{down}\in\R^{p\times r}, W_{up}\in\R^{r\times p}$ that constitute
$$x\longleftarrow x+\text{GeLU}(x\cdot W_{down})W_{up}$$
Hence the complexity, in comparison to full-finetuning, changes by replacing $d\to 2r$.

LoRA \cite{hu2021lora} also uses two matrices $W_{down}\in\R^{d\times r}, W_{up}\in\R^{r\times p}$ that constitute
$$x\longleftarrow x\cdot W+x\cdot W_{down}W_{up}$$
Hence the complexity, in comparison to full fine-tuning, changes by replacing $pd\to r(p+d)$.

\begin{table}[!htb]
\centering
\setlength{\tabcolsep}{1pt}
\caption{Per-layer time and space complexity of training on weights (full and parameter efficient fine-tuning) and biases. `$+$' means additional overhead to non-DP training.}
    \begin{tabular}{c||c||c|c|c|c||c|c}
         &forward&\multicolumn{4}{c||}{weight training}&\multicolumn{2}{|c}{bias training}\\
         \cline{3-8}
         & \&output grad&non-DP&DP full (Opacus)&DP LoRA&DP Adapter
         &non-DP&DP (ours)\\\hline
         \multirow{2}{*}{\shortstack{Time\\complexity}}&\multirow{2}{*}{$4BTpd$}&\multirow{2}{*}{$2BTpd$}&\multirow{2}{*}{$+2BTpd$}&\multirow{2}{*}{$+2BT(pr+dr)$}&\multirow{2}{*}{$+4BTpr$}
         &\multirow{2}{*}{$BTp$}&\multirow{2}{*}{$+3Bp$}  \\
         &&&&&&  \\\hline         \multirow{2}{*}{\shortstack{Space\\complexity}}&\multirow{2}{*}{$pd+BTd$}&\multirow{2}{*}{$BT(p+d)$}&\multirow{2}{*}{$+Bpd$}&\multirow{2}{*}{$+B(pr+dr)$}&\multirow{2}{*}{$+2Bpr$}
         &\multirow{2}{*}{$p$}&\multirow{2}{*}{$+Bp$}
         \\
         &&&&&&&
         \\\hline
         \# back-prop&&1&1&1&1&1&1
         \\\hline
forward hook &&\xmark&\cmark&\cmark&\cmark
&\xmark&\xmark
    \end{tabular}
    \label{tab:complexity overhead weight bias param eff}
\end{table}

For per-sample bias gradient clipping, we need $\frac{\partial \mathcal{L}_i}{\partial \b_{l}}^\top=\frac{\partial \mathcal{L}}{\partial \s_{l,i}}^\top\mathbf{1}$ in \Cref{eq:outer product linear}, which consists of the \textit{per-sample gradient instantiation} (i.e. summation along the feature dimension, from $\R^{Tp}\to\R^{p}$, $\frac{\partial \mathcal{L}}{\partial \s_{l,i}}\to\frac{\partial \mathcal{L}_i}{\partial \b_{l}}$), and computing the per-sample gradient norm (i.e. \textit{taking the square} at each index and \textit{summing all indices}). Here each operation in italic takes $Bp$ time complexity, meaning the total time complexity is $3Bp$, but the space complexity is $Bp$ if operated in-place.

\section{Experiment details}
\label{app:experiment details}
\subsection{Language tasks}
Throughout this work, the text datasets are processed and loaded from Huggingface \cite{lhoest-etal-2021-datasets}. We follow the same setup as \cite{li2021large,bu2022automatic}, such as $\delta=0.5/$sample size. The full fine-tuning is implemented by \href{https://github.com/lxuechen/private-transformers}{Private Transformers} codebase, version 0.2.0 (i.e. GhostClip algorithm \cite{li2021large}).

For text classification, we experiment on four datasets: \textbf{MNLI(m)}, the matched splits from Multi-Genre Natural Language Inference Corpus; \textbf{QQP}, the Quora Question Pairs2 dataset; \textbf{QNLI} The Stanford Question Answering dataset; \textbf{SST2} The Stanford Sentiment Treebank dataset.

To give a fair comparison, we use the same optimizer as in \cite{li2021large}, i.e. DP-Adam with Abadi's clipping.

\begin{table}[!htb]
    \centering
    \caption{Hyperparameters of text classification in \Cref{tab:finetuning comparison} and \Cref{tab:finetuning comparison eps3}, using RoBERTa (base/large).}
    \begin{tabular}{c|cccc}
     Dataset& MNLI&QQP&QNLI&SST2  \\\hline
 epoch&18&18&6&3\\\hline
 batch size&6000&6000&2000&1000\\\hline
 clipping threshold $R$&\multicolumn{4}{c}{0.1}\\\hline
 DP learning rate &\multicolumn{4}{c}{full 5e-4 / BiTFiT 5e-3}\\\hline
non-DP learning rate &\multicolumn{4}{c}{full 5e-5 / BiTFiT 1e-3}\\\hline
 max sequence length&\multicolumn{4}{c}{256}\\
    \end{tabular}
\end{table}

For E2E generation task, we experiment GPT2 models using the same optimizer as in \cite{bu2022automatic}, using DP-AdamW with automatic clipping.

\begin{table}[!htb]
    \centering
    \caption{Hyperparameters of E2E generation task in \Cref{tab:gpt e2e} and \Cref{tab:gpt e2e eps3}, using GPT2.}
    \begin{tabular}{c|ccc}
     Model& GPT2-small&GPT2-medium&GPT2-large  \\\hline
 epoch&\multicolumn{3}{|c}{10}\\\hline
 batch size&\multicolumn{3}{|c}{1024}\\\hline
 DP learning rate (full) &2e-3&2e-3&2e-3\\\hline
 non-DP learning rate (full) &2e-4&1e-4&1e-4\\\hline
  DP learning rate (BiTFiT) &\multicolumn{3}{|c}{1e-2}\\\hline
 non-DP learning rate (BiTFiT) &\multicolumn{3}{|c}{2e-3}\\\hline
 learning rate decay&\multicolumn{3}{|c}{No}\\\hline
 max sequence length&\multicolumn{3}{|c}{100}\\\hline
    \end{tabular}

\end{table}

\subsection{Image tasks}
\label{app:CV settings}
We give the experiments settings for image classification. For CIFAR10 and CIFAR100, we use the same setting as \cite{bu2022scalable}, e.g. 5 epochs for CrossViT, 3 epochs for ViT and BEiT-large. For CelebA, we use the same setting as \cite{bu2022automatic}, e.g. 10 epochs.

We use DP-Adam with Abadi's clipping. We do not apply tricks such as random data augmentation, weight standardization \cite{qiao2019micro}, or parameter averaging \cite{polyak1992acceleration}. Our experiments are heavily based on \href{https://github.com/woodyx218/private_CNN}{Private CNN} (i.e. MixGhostClip algorithm \cite{bu2022scalable}) and \href{https://github.com/rwightman/pytorch-image-models}{TIMM} codebases.

\begin{table}[!htb]
    \centering
    \caption{Hyperparameters of image classification task in \Cref{sec:image class},\Cref{tab:extended two-phase-cifar10},\Cref{tab:extended two-phase-cifar100},\Cref{tab:celebA 40}.}
    \begin{tabular}{c|cccc}
     Dataset& CIFAR10&CIFAR10&CIFAR100&CelebA  \\\hline
     Model& CrossViT&ViT-large&ViT-large&ResNet18  \\\hline
 epoch&5&3&3&10\\\hline
 batch size&1000&1000&1000&500\\\hline
clipping threshold&\multicolumn{4}{|c}{0.1}\\\hline
DP learning rate (full) &1e-3&5e-4&5e-4&1e-3\\\hline
DP learning rate (BiTFiT) &5e-3&5e-3&5e-3&8e-3\\\hline
learning rate decay&\multicolumn{4}{|c}{No}\\\hline
normalizing data&Yes&Yes&Yes&No\\\hline
   \end{tabular}
\end{table}

\clearpage
\section{Additional tables and figures}
\subsection{Parameter efficiency of DP-BiTFiT}
\begin{table}[!htb]
	\centering
	\setlength{\tabcolsep}{1pt}
\caption{Parameter efficiency of (DP) BiTFiT on various models.}
\begin{tabular}{c|c|c}
     Model&Number of params&\% of params \\\hline
VGG11&133M&0.009\\
VGG16& 138M&0.009\\
VGG19&144M&0.010\\
ResNet18& 11.7M&0.043\\
ResNet34& 21.8M&0.044\\
ResNet50& 25.6M&0.113\\
ResNet101& 44.5M&0.121\\
ResNet152& 60.2M&0.127\\
wide\_resnet50\_2&68.9M&0.051\\
wide\_resnet101\_2&126.9M&0.055\\
convnext\_base&88.6M&0.148\\
convnext\_large&197.8M&0.099\\
ViT-small-patch16& 22.0M&0.238\\
ViT-base-patch16& 86.6M&0.120\\
ViT-large-patch16& 304M&0.090\\
beit\_base\_patch16\_224& 86.5M&0.088\\
deit\_base\_patch16\_224& 86.4M&0.120\\
\hline
GPT2-small& 124M&0.082\\
GPT2-medium& 355M&0.076\\
GPT2-large&774M &0.066\\
RoBERTa-base& 125M&0.083\\
RoBERTa-large &355M &0.077\\
BERT-base-uncased& 109M&0.094\\
BERT-large-uncased& 335M&0.081\\
BART-large& 406M&0.082\\
longformer-base-4096&149M&0.088\\
longformer-large-4096&435M&0.080\\
\end{tabular}

\label{tab:parameter efficiency extended}
\end{table}

\subsection{More results on DP-BiTFiT and language tasks}

\begin{table}[!htb]
\centering
	\setlength{\tabcolsep}{1pt}
	\caption{Accuracy of full fine-tuning and BiTFiT with RoBERTa, under different per-sample clipping functions (indicated as subscript, Abadi \cite{abadi2016deep} and AUTO-S \cite{bu2022automatic}). Same setting as \Cref{app:experiment details}.}
		\begin{tabular}{|c||c|c|c|c|c|c|c|c|c|c|}
	    \hline
		& \multicolumn{5}{|c|}{full \citep{li2021large,bu2022automatic}}&\multicolumn{5}{|c|}{BiTFiT (ours)}
		\\\hline
			\multicolumn{11}{|c|}{RoBERTa-base}
			\\\hline
			&\textcolor{cyan}{standard}&DP$_\text{Abadi}$&DP$_\text{AUTO}$&DP$_\text{Abadi}$&DP$_\text{AUTO}$&\textcolor{cyan}{standard}&DP$_\text{Abadi}$&DP$_\text{AUTO}$&DP$_\text{Abadi}$&DP$_\text{AUTO}$
			\\
			&\textcolor{cyan}{$\epsilon=\infty$}&$\epsilon=8$&$\epsilon=8$&$\epsilon=3$&$\epsilon=3$&\textcolor{cyan}{$\epsilon=\infty$}&$\epsilon=8$&$\epsilon=8$&$\epsilon=3$&$\epsilon=3$
			\\\hline
			Accuracy SST2&\textcolor{cyan}{94.5}&92.1&92.4&91.9&92.3&\textcolor{cyan}{93.5}&92.4&92.4&92.2&92.2
			\\
			Accuracy QNLI&\textcolor{cyan}{91.4}&87.9&87.9&87.4&86.9&\textcolor{cyan}{87.3}&86.9&87.0&86.4&86.4
			\\
			Accuracy QQP&\textcolor{cyan}{87.3}&86.1&86.6&85.6&85.8&\textcolor{cyan}{86.1}&85.6&85.9&84.8&85.0
			\\
			Accuracy MNLI-m&\textcolor{cyan}{85.9}&83.2&83.8&82.5&83.2&\textcolor{cyan}{83.4}&82.9&83.2&82.5&82.7
			\\\hline\hline
		\multicolumn{11}{|c|}{RoBERTa-large}
			\\\hline
			&\textcolor{cyan}{standard}&DP$_\text{Abadi}$&DP$_\text{AUTO}$&DP$_\text{Abadi}$&DP$_\text{AUTO}$&\textcolor{cyan}{standard}&DP$_\text{Abadi}$&DP$_\text{AUTO}$&DP$_\text{Abadi}$&DP$_\text{AUTO}$
			\\
			&\textcolor{cyan}{$\epsilon=\infty$}&$\epsilon=8$&$\epsilon=8$&$\epsilon=3$&$\epsilon=3$&\textcolor{cyan}{$\epsilon=\infty$}&$\epsilon=8$&$\epsilon=8$&$\epsilon=3$&$\epsilon=3$
			\\\hline
			Accuracy SST2&\textcolor{cyan}{96.2}&93.8&94.6&93.0&93.9&\textcolor{cyan}{95.5}&94.5&94.7&94.5&94.6
			\\
			Accuracy QNLI&\textcolor{cyan}{93.6}&91.1&91.5&90.8&91.0&\textcolor{cyan}{92.2}&91.1&91.3&90.7&90.8
			\\
			Accuracy QQP&\textcolor{cyan}{87.9}&86.9&87.5&86.6&86.8&\textcolor{cyan}{87.9}&86.9&87.1&86.6&86.7
			\\
			Accuracy MNLI-m&\textcolor{cyan}{90.3}&87.0&87.1&86.4&86.3&\textcolor{cyan}{89.3}&88.3&88.4&87.2&87.8
			\\\hline
		\end{tabular}

	\label{tab:finetuning comparison eps3}
\end{table}

\newpage
\begin{table}[!htb]
	\centering
	\caption{Accuracy of fine-tuning with GPT2 on E2E dataset. LoRA and prefix results are taken from \cite{li2021large}. Same setting as \Cref{app:experiment details}.}
 \resizebox{\linewidth}{!}{
		\begin{tabular}{c|c|c|c|c|c|c|c|c|c}
			Model&Fine-tuning&\% of params&Privacy$\downarrow$&Perplexity$\downarrow$&BLEU$\uparrow$&ROGUE-L$\uparrow$&NIST$\uparrow$&METEOR$\uparrow$&CIDEr$\uparrow$\\\hline
			\multirow{8}{*}{\shortstack{GPT2-small\\(124M)}}&\multirow{2}{*}{full}&\multirow{2}{*}{100\%}&\textcolor{cyan}{standard}&\textcolor{cyan}{2.91}&\textcolor{cyan}{69.46} &\textcolor{cyan}{71.36} &\textcolor{cyan}{8.78} &\textcolor{cyan}{0.46}&\textcolor{cyan}{2.42}\\
			&&&DP ($\epsilon=8$)&2.33&63.60&67.07&7.71&0.40&1.94\\
			&&&DP ($\epsilon=3$)&2.36&61.34&65.87&7.07&0.39&1.80\\\cline{2-10}
			&\multirow{2}{*}{LoRA}&\multirow{2}{*}{---}&\textcolor{cyan}{standard}&\textcolor{cyan}{---}&\textcolor{cyan}{69.68} &\textcolor{cyan}{71.71} &\textcolor{cyan}{8.82} &\textcolor{cyan}{0.46}&\textcolor{cyan}{2.49}\\
			&&&DP ($\epsilon=8$)&---&63.39&67.53&7.45&0.41&1.95\\
			&&&DP ($\epsilon=3$)&---&58.15&65.77&5.46&0.37&1.58\\\cline{2-10}
			&\multirow{2}{*}{prefix}&\multirow{2}{*}{---}&\textcolor{cyan}{standard}&\textcolor{cyan}{---}&\textcolor{cyan}{68.85} &\textcolor{cyan}{70.81} &\textcolor{cyan}{8.72} &\textcolor{cyan}{0.45}&\textcolor{cyan}{2.35}\\
			&&&DP ($\epsilon=8$)&---&49.26&60.73&5.53&0.36&1.57\\
			&&&DP ($\epsilon=3$)&---&47.77&58.96&5.25&0.36&1.51\\\cline{2-10}
			&\multirow{2}{*}{BiTFiT}&\multirow{2}{*}{0.082\%}&\textcolor{cyan}{standard}&\textcolor{cyan}{3.19}&\textcolor{cyan}{64.46} &\textcolor{cyan}{63.67} &\textcolor{cyan}{4.25} &\textcolor{cyan}{0.36}&\textcolor{cyan}{1.36}\\
			&&&DP ($\epsilon=8$)&2.89&60.56&64.96&6.14&0.37&1.62\\
			&&&DP ($\epsilon=3$)&3.00&54.78&63.55&4.78&0.34&1.31\\\hline\hline
			\multirow{4}{*}{\shortstack{GPT2-medium\\(355M)}}&\multirow{2}{*}{full}&\multirow{2}{*}{100\%}&\textcolor{cyan}{standard}&\textcolor{cyan}{2.08}&\textcolor{cyan}{68.50} &\textcolor{cyan}{71.46} &\textcolor{cyan}{8.63} &\textcolor{cyan}{0.45}&\textcolor{cyan}{2.14}\\
			&&&DP ($\epsilon=8$)&2.25&64.22&67.53&8.17&0.42&2.08\\
			&&&DP ($\epsilon=3$)&2.62&63.85&67.07&7.11&0.39&1.75\\\cline{2-10}
			&\multirow{2}{*}{BiTFiT}&\multirow{2}{*}{0.076\%}&\textcolor{cyan}{standard}&\textcolor{cyan}{2.85}&\textcolor{cyan}{64.48} &\textcolor{cyan}{67.81} &\textcolor{cyan}{8.50} &\textcolor{cyan}{0.43}&\textcolor{cyan}{2.11}\\
			&&&DP ($\epsilon=8$)&2.67&61.02&66.13&7.18&0.39&1.80\\
			&&&DP ($\epsilon=3$)&2.67&57.11&66.16&5.07&0.37&1.47\\\hline\hline
			\multirow{4}{*}{\shortstack{GPT2-large\\(774M)}}&\multirow{2}{*}{full}&\multirow{2}{*}{100\%}&\textcolor{cyan}{standard}&\textcolor{cyan}{1.79}&\textcolor{cyan}{66.84} &\textcolor{cyan}{70.38} &\textcolor{cyan}{8.73} &\textcolor{cyan}{0.46}&\textcolor{cyan}{2.36}\\
			&&&DP ($\epsilon=8$)&2.26&64.64&68.97&8.30&0.42&2.16\\
			&&&DP ($\epsilon=3$)&2.65&64.18&67.86&7.94&0.40&2.01\\\cline{2-10}
			&\multirow{2}{*}{BiTFiT}&\multirow{2}{*}{0.066\%}&\textcolor{cyan}{standard}&\textcolor{cyan}{2.79}&\textcolor{cyan}{65.79} &\textcolor{cyan}{67.61} &\textcolor{cyan}{8.55} &\textcolor{cyan}{0.43}&\textcolor{cyan}{2.21}\\
	&&&DP ($\epsilon=8$)&2.59&65.21&67.88&8.43&0.42&2.15\\
			&&&DP ($\epsilon=3$)&2.61&65.18&67.90&8.34&0.42&2.12\\\hline
	\end{tabular}
	}

	\label{tab:gpt e2e eps3}
\end{table}

\subsection{More results on two-phase training}
\label{app:more results image}
Here X$+$BiTFiT does not train last layer, i.e. the classification head is randomized before full fine-tuning happens.

\begin{table}[!htb]
\centering
\setlength{\tabcolsep}{1pt}
\caption{Accuracy of two-phase fine-tuning on CIFAR10. Same setting as \Cref{app:CV settings}. BEiT-large uses DP full fine-tuning learning rate 5e-4, DP-BiTFiT learning rate 5e-3. Others use DP full fine-tuning learning rate 1e-3, DP-BiTFiT learning rate 5e-3.}
\begin{tabular}{|c|c|c|c|c|c|c|c|}
\hline
\multicolumn{6}{|c|}{CIFAR10}
\\\hline
Model&Privacy&0+BiTFiT&1+BiTFiT&2+BiTFiT&DP full
\\\hline
beit\_large\_patch16\_224&$\epsilon=1$&11.7&98.2&97.9&97.2
\\
&$\epsilon=2$&10.0&98.3&98.0&97.3
\\
&$\epsilon=4$&13.8&98.2&98.0&97.5
\\
&$\epsilon=8$&10.1&98.5&98.0&97.8
\\\hline
beit\_base\_patch16\_224&$\epsilon=1$&10.0&96.6&96.0&95.4
\\
&$\epsilon=2$&10.7&97.1&96.4&96.0
\\
&$\epsilon=4$&14.0&97.2&96.6&96.2
\\
&$\epsilon=8$&10.0&97.2&96.5&96.3
\\\hline
deit\_base\_patch16\_224&$\epsilon=1$&78.2&94.4&95.2&95.4
\\
&$\epsilon=2$&75.0&95.4&95.2&95.6
\\
&$\epsilon=4$&72.9&95.8&95.9&96.0
\\
&$\epsilon=8$&71.2&96.1&96.0&96.3
\\\hline
crossvit\_base\_240&$\epsilon=1$&74.3&92.4&94.3&95.2
\\
&$\epsilon=2$&80.4&93.6&95.0&95.3
\\
&$\epsilon=4$&81.0&94.9&95.8&95.7
\\
&$\epsilon=8$&78.2&94.8&95.8&96.2
\\\hline
vit\_large\_patch16\_224&$\epsilon=1$&89.7&98.9&98.7&98.9
\\
&$\epsilon=2$&90.6&98.8&98.9&98.9
\\
&$\epsilon=4$&93.2&98.9&98.8&99.0
\\
&$\epsilon=8$&93.9&99.0&98.9&99.0
\\\hline
vit\_base\_patch16\_224&$\epsilon=1$&86.7&95.2&97.0&96.8
\\
&$\epsilon=2$&89.3&97.7&97.1&97.1
\\
&$\epsilon=4$&88.3&97.7&97.2&97.2
\\
&$\epsilon=8$&88.7&97.6&97.2&97.4
\\\hline
\end{tabular}

\label{tab:extended two-phase-cifar10}
\end{table}

\begin{table}[!htb]
\centering
\setlength{\tabcolsep}{1pt}
\caption{Accuracy of two-phase fine-tuning on CIFAR100. Same setting as \Cref{app:CV settings}. BEiT-large uses DP full fine-tuning learning rate 5e-4, DP-BiTFiT learning rate 5e-3. Others use DP full fine-tuning learning rate 1e-3, DP-BiTFiT learning rate 5e-3.}
\begin{tabular}{|c|c|c|c|c|c|c|c|}
\hline
\multicolumn{6}{|c|}{CIFAR100}
\\\hline
Model&Privacy&0+BiTFiT&1+BiTFiT&2+BiTFiT&DP full
\\\hline
beit\_large\_patch16\_224&$\epsilon=1$&1.0&86.9&87.8&87.0
\\
&$\epsilon=2$&1.0&88.7&89.3&88.7
\\
&$\epsilon=4$&1.0&89.7&89.7&89.6
\\
&$\epsilon=8$&1.0&90.3&90.7&90.0
\\\hline
beit\_base\_patch16\_224&$\epsilon=1$&1.0&81.4&82.2&80.9
\\
&$\epsilon=2$&1.0&83.4&83.4&83.1
\\
&$\epsilon=4$&1.0&84.6&85.1&84.8
\\
&$\epsilon=8$&1.0&84.9&85.6&85.2
\\\hline
deit\_base\_patch16\_224&$\epsilon=1$&10.9&49.1&65.9&69.1
\\
&$\epsilon=2$&13.6&58.1&71.5&74.3
\\
&$\epsilon=4$&15.7&64.5&73.9&77.1
\\
&$\epsilon=8$&16.6&69.7&75.7&77.9
\\\hline
crossvit\_base\_240&$\epsilon=1$&12.2&49.2&61.7&67.6
\\
&$\epsilon=2$&12.3&56.8&65.3&71.6
\\
&$\epsilon=4$&17.2&61.6&70.4&73.1
\\
&$\epsilon=8$&20.9&63.4&72.8&74.2
\\\hline
vit\_large\_patch16\_224&$\epsilon=1$&14.0&73.5&86.0&87.7
\\
&$\epsilon=2$&19.4&82.4&89.0&90.1
\\
&$\epsilon=4$&24.3&87.5&89.9&91.0
\\
&$\epsilon=8$&23.9&89.0&90.7&91.3
\\\hline
vit\_base\_patch16\_224&$\epsilon=1$&16.0&64.3&79.5&83.9
\\
&$\epsilon=2$&22.9&77.0&83.8&85.5
\\
&$\epsilon=4$&21.2&83.0&85.2&87.2
\\
&$\epsilon=8$&26.2&83.8&86.5&87.1
\\\hline
\end{tabular}
\label{tab:extended two-phase-cifar100}
\end{table}

\clearpage
\begin{table}[!htb]
    \centering
\caption{Accuracy on CelebA dataset with settings in \Cref{app:CV settings} from one run. DP full fine-tuning is implemented with the most efficient MixGhostClip algorithm \cite{bu2022scalable}. We observe that linear probing (LP) only gives 83.67\% at $\epsilon=8$. *Note the accuracy is based on \texttt{timm<=0.6.5} and may change for a different version.}
\resizebox{\linewidth}{!}{
    \begin{tabular}{|c||c|c|c|c|c||c|c|c|c|c|}
\hline
\multirow{2}{*}{Attributes}&0+BiTFiT&1+BiTFiT&2+BiTFiT&DP full&DP-BiTFiT(LP)&0+BiTFiT&1+BiTFiT&2+BiTFiT&DP full&DP-BiTFiT(LP)\\\cline{2-11}
&\multicolumn{5}{|c||}{$\epsilon=3$}&\multicolumn{5}{|c|}{$\epsilon=8$}\\\hline
5 o Clock Shadow&90.01&90.01&90.14&91.32&90.35
&90.01&90.01&90.51&91.64&90.97\\
Arched Eyebrows&71.56&73.12&76.01&77.33&75.41
&71.56&73.74&75.49& 78.82&76.49\\
Attractive&68.71&73.98&75.99&79.22&74.96
&69.70&73.61&76.20& 78.08&7523\\
Bags Under Eyes&79.74&79.76&81.27&81.73&81.14
&79.74&79.74&80.69& 82.62&8172\\
Bald&97.88&97.88&97.88&97.93&97.93
&97.88&97.88&97.88&97.91&9790\\
Bangs&84.43&84.43&84.80&94.06&90.85
&84.43&84.44&86.51&94.22&92.34\\
Big Lips&67.30&67.30&67.30&67.78&67.42
&67.30&67.30&67.29& 68.34&67.65\\
Big Nose&78.80&78.95&80.08&81.19&79.96
&78.80&78.92&79.23&81.86&80.28\\
Black Hair&72.84&74.86&82.37&85.84&81.48
&73.02&78.71&83.33& 86.47&82.38\\
Blond Hair&89.54&93.00&93.28&94.17&93.03
&89.13&92.62&93.88&94.34&93.51\\
Blurry&94.94&94.94&94.94&95.05&95.21
&94.94&94.94&94.96&95.10&95.34\\
Brown Hair&82.03&82.02&82.87&85.44&82.68
&82.03&82.37&83.49& 85.04&82.88\\
Bushy Eyebrows&87.05&87.05&87.21&88.26&87.11
&87.05&87.05&87.15& 89.02&87.22\\
Chubby&94.70&94.70&94.70&94.84&94.57
&94.70&94.70&94.70& 94.78&94.47\\
Double Chin&95.43&95.43&95.43&95.49&95.34
&95.43&95.43&95.43& 95.39&95.26\\
Eyeglasses&93.54&93.54&93.54&94.30&94.77
&93.54&93.54&93.54& 95.85&96.32\\
Goatee&95.42&95.42&95.42&95.96&95.41
&95.42&95.42&95.42& 95.89&95.55\\
Gray Hair&96.81&96.81&96.85&97.44&96.78
&96.81&96.81&97.12& 97.45&96.59\\
Heavy Makeup&76.51&82.76&85.71&88.48&83.73
&77.22&83.03&85.86& 89.05&84.70\\
High Cheekbones&62.13&68.20&81.63&83.77&76.91
&61.43&67.27&81.33& 84.20&79.42\\
Male&80.37&88.47&91.52&94.73&89.92
&82.04&88.52&92.14& 95.19&90.69\\
Mouth Slightly Open&54.03&59.32&77.61&86.75&74.20
&55.26&60.70&79.42& 90.24&77.53\\
Mustache&96.13&96.13&96.13&96.10&96.06
&96.13&96.13&96.13& 96.12&95.98\\
Narrow Eyes&85.13&85.13&85.13&85.14&85.15
&85.13&85.13&85.13& 85.16&85.13\\
No Beard&85.37&85.87&87.56&92.94&88.33
&85.37&85.88&88.59& 93.59&89.81\\
Oval Face&70.44&70.94&71.50&73.11&71.51
&70.44&71.48&71.92& 71.77&71.25\\
Pale Skin&95.79&95.79&95.79&95.79&95.76
&95.79&95.79&95.79& 95.79&95.73\\
Pointy Nose&71.43&71.51&71.63&71.89&71.40
&71.43&71.47&71.77& 72.87&72.11\\
Receding Hairline&91.51&91.51&91.51&91.59&91.40
&91.51&91.51&91.51& 91.61&91.39\\
Rosy Cheeks&92.83&92.83&92.86&93.07&92.75
&92.87&92.83&92.86& 93.33&92.99\\
Sideburns&95.36&95.36&95.36&96.44&95.55
&95.36&95.36&95.36& 96.63&95.79\\
Smiling&60.07&66.32&85.85&89.34&79.99
&58.92&65.97&85.55& 89.11&82.82\\
Straight Hair&79.01&79.01&79.02&79.65&79.22
&79.01&79.01&79.13& 78.60&79.47\\
Wavy Hair&71.24&73.09&76.22&77.35&77.98
&70.86&73.62&77.11& 72.73&78.90\\
Wearing Earrings&79.34&79.34&80.37&83.24&81.54
&79.34&79.34&80.71& 84.36&82.65\\
Wearing Hat&95.80&95.80&95.80&96.01&95.95
&95.80&95.80&95.80& 97.02&96.63\\
Wearing Lipstick&80.61&87.90&89.81&91.59&87.54
&80.35&87.20&89.56& 91.94&88.16\\
Wearing Necklace&86.21&86.21&86.21&86.21&86.16
&86.21&86.21&86.21& 86.21&86.12\\
Wearing Necktie&92.99&92.99&93.03&93.58&93.61
&92.99&92.99&93.11& 93.57&94.13\\
Young&75.71&79.33&81.23&83.69&80.57
&75.71&78.52&80.66& 83.11&80.93\\\hline
Average&82.97&84.42&86.54&88.20&\textcolor{red}{86.25}
&83.01&84.52&86.71&88.38&\textcolor{red}{86.87}\\\hline
Total time&10:30&12:02&13:34&25:50&\textcolor{red}{10:30}
&10:30&12:02&13:34&25:50&\textcolor{red}{10:30}\\\hline
\end{tabular}
}
\label{tab:celebA 40}
\end{table}

\subsection{Hyperparameter tuning for DP-BiTFiT}
We demonstrate that employing DP-BiTFiT does not complicate the learning rate tuning, when compared to the full fine-tuning.
\begin{table}[!htb]
\centering
    \vspace{-0.3cm}
    \caption{Test accuracy on SST2 under $\epsilon=8$, using DP-Adam with AUTO-S clipping.}
	\setlength{\tabcolsep}{3pt}
    \begin{tabular}{c||c|c|c|c|c|c|c|c|c|c|c|c|c}
    &\multicolumn{5}{|c|}{DP-BiTFiT}&\multicolumn{4}{|c|}{DP full}&\multicolumn{4}{|c}{non-DP full}\\\hline
    learning rate&5e-4&1e-3&2e-3&5e-3&1e-2
    &1e-4&2e-4&5e-4&1e-3
    &1e-5&2e-5&5e-5&1e-4\\\hline
 RoBERTa-base&  90.94&91.28&91.74&92.43&90.94
 &91.51&91.97&92.43&91.28
 &93.92&94.38&94.49&93.35\\
 RoBERTa-large& 94.38&95.07&94.38&94.50&94.04
 &94.84&94.72&94.61&92.66
 &95.76 &96.21 &96.21 &95.99
    \end{tabular}
    \label{tab:tuning learning rate}
\vspace{-0.5cm}
\end{table}

\end{document}